\documentclass[single-column]{fairmeta}

\usepackage[table]{xcolor}

\definecolor{StanfordRed}{HTML}{8C1515}
\definecolor{StanfordDark}{HTML}{2E2D29}
\definecolor{StanfordGray}{HTML}{53565A}

\definecolor{NTURed}{HTML}{D71440}
\definecolor{NTUBlue}{HTML}{181C62}

\colorlet{ReportPrimary}{NTUBlue}
\colorlet{ReportAccent}{NTURed}
\colorlet{ReportSecondary}{StanfordRed}
\colorlet{ReportText}{StanfordDark}
\colorlet{ReportGray}{StanfordGray}

\colorlet{SoftBlue}{NTUBlue!6}
\colorlet{SoftRed}{NTURed!6}
\colorlet{SoftStanford}{StanfordRed!6}
\colorlet{SoftGray}{StanfordGray!7}

\colorlet{BlueAccent}{NTUBlue!72!white}
\colorlet{RedAccent}{NTURed!78!StanfordRed}
\colorlet{PurpleAccent}{NTURed!45!NTUBlue}

\definecolor{newyellow}{HTML}{FFD94D}
\definecolor{newgrey}{HTML}{7F7F7F}
\definecolor{newpink}{HTML}{FBCDF4}

\definecolor{realworldoft}{HTML}{029533}
\definecolor{realworldsf}{HTML}{8CC46A}
\definecolor{realworldour}{HTML}{FFC715}

\usepackage{amsmath,amssymb,mathtools,amsthm}
\usepackage{siunitx,nicefrac}

\usepackage{booktabs,array,tabularx,makecell,multirow}

\usepackage{graphicx,float,wrapfig}

\usepackage{algorithm}
\usepackage{algpseudocode}

\usepackage{microtype,xspace,enumitem,listings,utfsym}

\usepackage[most]{tcolorbox}
\tcbuselibrary{theorems}

\usepackage{url}
\usepackage{hyperref}

\newtcbtheorem{finding}{Finding}{
    colback=SoftBlue,
    colframe=ReportPrimary,
    coltitle=white,
    fonttitle=\bfseries,
    boxrule=0.8pt,
    arc=2pt,
}{find}

\newtcolorbox{keypoint}{
    colback=SoftRed,
    colframe=ReportAccent,
    fonttitle=\bfseries,
    title=Key Point,
    boxrule=0.8pt,
    arc=2pt,
}

\newtcolorbox{reportnote}{
    colback=SoftGray,
    colframe=ReportGray,
    fonttitle=\bfseries,
    title=Note,
    boxrule=0.6pt,
    arc=2pt,
}

\title{RoboFoundry: System-as-Policy Evolution for Self-Learning Embodied Agents}

\author[1,*]{Jingsong Liang}
\author[1,2,*]{Shuhao Liao}
\author[1]{Shizhe Zhang}
\author[2]{Diyuan Hou}
\author[1]{Yuxin Cai}
\author[3]{Xinjian Deng}
\author[3]{Chengyang He}
\author[1]{Wenhui Huang}
\author[1]{Runjia Tan}
\author[1]{Zhidong Wang}
\author[4]{Lan Yu}
\author[4]{Xuesong Tian}
\author[3]{Guillaume Sartoretti}
\author[2]{Jie Luo}
\author[5]{Yao Mu}
\author[2,\ddagger]{Wenjun Wu}
\author[1,\ddagger]{Wanhua Li}
\author[1,\ddagger]{Chen Lv}

\affiliation[1]{{\footnotesize Nanyang Technological University}}
\affiliation[2]{{\footnotesize Beihang University}}
\affiliation[3]{{\footnotesize National University of Singapore}}
\affiliation[4]{{\footnotesize Cloud Butterfly Technology}}
\affiliation[5]{{\footnotesize Shanghai Jiao Tong University}}

\contribution[*]{{\footnotesize Equal Contribution}}
\contribution[\ddagger]{{\footnotesize Corresponding Author}}

\metadata[Website]{%
  \href{https://jingsongliang.com/robofoundry}%
       {jingsongliang.com/robofoundry}%
}

\abstract{
A foundation model should not act in isolation as an embodied agent. Yet, existing methods often optimize individual components of the agent stack, such as memory, context, skills, or action interfaces, rather than treating the supporting system itself as a unified policy. Moreover, interaction alone does not yield self-improvement unless execution experience is converted into persistent, validated system changes. We therefore propose \textbf{RoboFoundry}, the first embodied agentic framework that formulates this process as \textbf{Self-Evolving System-as-Policy}. RoboFoundry diagnoses capability gaps in \emph{decision-making} and \emph{memory management}, converts execution traces into validated task-specific system updates, and promotes recurring improvements to the general system. Evolution operates over two complementary surfaces: a context system that manages active internal context and persistent file-system memory, and a hierarchical skill system that organizes atomic skills, reusable compositions, and failure-conditioned recovery. A shared semantic interface separates embodiment-invariant decisions from embodiment-specific execution, allowing evolved system capabilities to transfer across heterogeneous robots. On EmbodiedBench, RoboFoundry achieves state-of-the-art performance, notably improving GPT-5.5 by \SI{27.8}{\%}. It also brings Qwen3.7-Plus to near parity with GPT-5.5 (\SI{70.3}{\%} vs. \SI{72.7}{\%}), showing consistent gains from system-as-policy evolution across foundation models. For long-horizon memory, RoboFoundry outperforms all baselines on RoboMemArena by at least \SI{39.0}{\%}, even against methods assisted by external foundation models. On LIBERO-PRO, it further outperforms Cap-Agent0 by \SI{243.8}{}--\SI{679.7}{\%} across all perturbation types.
In real-world deployments, RoboFoundry demonstrates \textit{zero-shot transfer} and \textit{online evolution} across robots and tasks, highlighting its potential for fully autonomous embodied agents.
}

\begin{document}

\maketitle

\section{Introduction}
\label{sec:introduction}

Recent advances in large language models (LLMs) and multimodal large
language models (MLLMs) are shifting embodied intelligence from learning a
separate policy per task toward using general-purpose foundation models as
embodied decision makers.
Early work mainly uses them as high-level planners, exposing observations,
robot states, and admissible actions through designed prompts
\citep{brohan2023can,pmlr-v205-huang23c,pmlr-v202-jiang23b}.
As model outputs become increasingly executable, \textit{code-as-policy}
methods take the next step, composing perception and control primitives into
programs that interact with the environment
\citep{fu2026capx,liu2026guava,li2026roboclaw,zhang2026playful}, moving the
foundation model from a planner inside the robotic stack toward the decision
core of a closed-loop embodied agent. 
\begin{figure}[t]
\centering
\includegraphics[width=0.98\textwidth]
{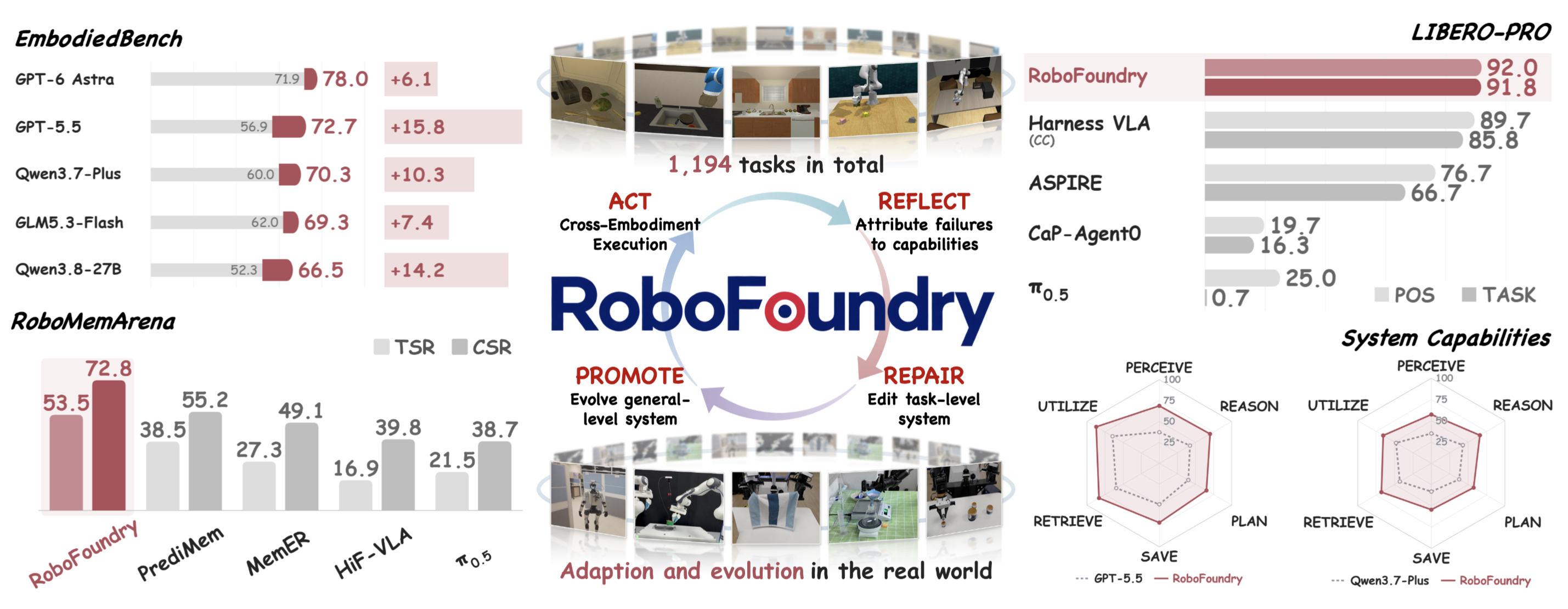}
\caption{
\textbf{RoboFoundry: System-as-Policy Evolution across benchmarks and real robots.}
The central Act--Reflect--Repair--Promote loop executes across embodiments,
attributes failures to capabilities, revises the task-level system, and
promotes validated changes to the general system.
The surrounding panels show gains across foundation models on
EmbodiedBench, long-horizon memory results on RoboMemArena, and robustness perturbations
under LIBERO-PRO.
The capability plots compare GPT-5.5 and Qwen3.7-Plus before and after
RoboFoundry, while the lower image strip illustrates adaptation and
evolution on physical robots.
}
\label{fig:robo_teaser}
\end{figure}

\textbf{Yet a foundation model alone does not define such an agent.}
The same model behaves very differently as the surrounding interaction
structure changes: iterative observation, feedback, and verification support
better-grounded decisions, while structured action interfaces reduce the
burden of turning semantic decisions into executable behavior
\citep{li2024embodied,liu2026guava,kwok2026llm}.
\emph{The executable embodied policy is therefore a system, rather than a
model in isolation}, jointly determined by the model and by what information
reaches it, what experience persists, and how its decisions become physical
actions.
Existing embodied agents, however, are rarely optimized as such a system, and
instead refine a predefined component while leaving the rest unchanged:
interaction harnesses improve model--environment feedback
\citep{liu2026guava}; memory systems improve access to past information
\citep{huang2026roboharness,zhang2026harness}; skill-learning agents
accumulate reusable behaviors
\citep{wang2023voyager,lu2026aspire,wang2026skillmemo}; and coding agents
refine programs from physical feedback \citep{xiao2026enpire}.
Such gains can be substantial, but remain \emph{component-local}: improving
one component neither identifies another bottleneck nor determines how the
improvement propagates through the system.
\textbf{Richer interaction also does not by itself make an embodied system
self-improving.}
Most systems stay fixed after deployment even though interaction continuously
exposes where their support is insufficient, so the same weakness recurs
across episodes
\citep{pmlr-v205-huang23c,madaan2023self,brohan2023can,zhang2026harness};
The resulting experience is typically used for evaluation or manual debugging
rather than converted into persistent, validated system changes
\citep{li2024embodied,yang2025embodiedbench,fu2026capx,liu2026guava}.

These observations lead to the central question:
\textbf{Can an embodied agent treat its entire system as the policy, diagnose
where that system limits behavior, and evolve the corresponding components
from embodied experience?}
To this end, we propose \textbf{RoboFoundry}, the first embodied agent
framework that formulates the agent stack itself as a
\textbf{Self-Evolving System-as-Policy}.
Rather than committing in advance to one module to optimize, RoboFoundry lets
embodied execution history decide which part of the system supporting the
foundation model should change and how far that change should propagate,
extending the progression from \emph{language-model planning} to
\emph{code-as-policy execution}, and further to
\emph{system-as-policy evolution}.

Evolution operates over two complementary \emph{surfaces}.
The \emph{context surface} governs how observations and accumulated
experience are managed, spanning active internal context within an episode
and persistent file-system memory across episodes.
The \emph{skill surface} organizes executable behavior, from atomic skills to
reusable compositions and failure-conditioned recovery graphs. Execution traces decide both \emph{why} the system is limited and
\emph{where} it should change: RoboFoundry attributes recurring failures to
model-conditioned capability gaps, confines each revision to the responsible
surface, and assigns it an explicit \emph{scope}, where a task-specific
repair is validated in place while a general update is promoted only after
the same improvement proves effective beyond the held-in task.
Self-evolution is therefore a controlled loop of diagnosis, intervention,
validation, and promotion, rather than unconstrained self-rewriting.

A shared semantic interface keeps evolved capability from being bound to one
robot or one backbone: decisions are expressed once as semantic units, while
replaceable embodiment bindings realize them through robot-specific
primitives.
Because evolution acts on the semantic side, a system evolved on one platform
is reused on another by swapping its bindings, and the same evolved support
attaches to different foundation models.

Multiple embodied benchmarks test whether system-level evolution improves
decision-making, long-horizon memory, and robustness under task and
environment perturbations, repeated across foundation models of different
capability to verify that the gains come from the system rather than from a
particular backbone; real-robot deployments then test transfer to unseen
robots and tasks and continued improvement after deployment.
RoboFoundry achieves state-of-the-art results in all settings and keeps
evolving across simulation and the real world.

In summary, we make key contributions as follows:
\begin{itemize}
    \item We propose \textbf{RoboFoundry}, the first embodied agent framework
    that formulates the agent system itself as the policy and makes it
    \textit{self-evolving}, advancing embodied agents from
    \emph{code-as-policy execution} to \textbf{system-as-policy evolution}.

    \item We develop a capability-guided mechanism for context--skill
    co-evolution that turns execution traces into targeted, validated system
    revisions and promotes recurring improvements from task-specific support
    to the general system.

    \item We introduce a shared semantic interface that decouples
    embodiment-invariant decisions from embodiment-specific execution, making
    evolved capability reusable across heterogeneous robots and tasks.

    \item RoboFoundry delivers \textbf{substantial gains across multiple
    benchmarks}, spanning embodied decision-making, long-horizon memory, and
    robustness to perturbations, brings open-source backbones close to
    frontier-model performance, and demonstrates \textbf{zero-shot
    generalization and online evolution} in real-world robotic experiments.
\end{itemize}

\section{Method}
\label{sec:method}

\subsection{Self-Evolving System as Policy}
\label{sec:system_policy}

RoboFoundry treats the overall system supporting task execution as the
policy to optimize.
Let $M$ denote a frozen foundation model and $H(M)$ its supporting system.
Rather than updating the parameters of $M$, RoboFoundry improves the
effective capability of the model--system pair by evolving $H(M)$ through
filesystem operations.
The model inspects the supporting system using operations such as
\texttt{cat} and \texttt{grep}, and edits it using \texttt{add},
\texttt{modify}, and \texttt{delete}.
We decompose the supporting system into a task-specific component $H_t$
and a general component $H_g$, with corresponding optimization spaces
$S_t$ and $S_g$.
The task-specific space $S_t$ comprises semantic-level task formulation
$S_t^c$ and embodiment-specific execution $S_t^e$.

RoboFoundry organizes system evolution as an inner--outer loop.
The \textbf{inner loop} executes tasks under the current system and collects
execution traces.
For a task instance $x\sim\mathcal{X}$, a rollout is written as
$\tau\sim\mathcal{T}(H_g,H_t,x)$, where $\tau$ records system and tool calls,
step-wise observations and robot states, execution outcomes, and task feedback.
Here, $\mathcal{T}(H_g,H_t,x)$ denotes the set of stored traces.
The \textbf{outer loop} proposes system edits, which are deployed and
evaluated through subsequent inner-loop rollouts.
We detail task execution in Section~\ref{sec:inner_loop}.

The outer loop first performs \emph{task-level system improvement} over
$S_t$ to improve completion of the current task.
Here, $r_x(\Delta H_t, \tau)$ denotes the performance score of rollout $\tau$
on task $x$, where $\Delta H_t$ is a candidate edit to the
task-level system.
Based on execution traces, the outer loop revises either the
semantic-level system in $S_t^c$ or the embodiment-specific execution
system in $S_t^e$:
\begin{equation}
H_t^{*}
=
\arg\max_{H_t \in S_t}
\mathbb{E}_{\tau}
\left[
r_x(\Delta H_t,\tau)
\right].
\label{eq:task_level_objective}
\end{equation}
Beyond the current task, RoboFoundry performs \emph{general-level system
improvement} over the persistent system $H_g$.
It examines whether a successful task-level improvement addresses a
capability gap shared by other tasks, using traces accumulated across
tasks and embodiments.
Let $\mathcal{X}_{\mathcal{A}}$ denote the task distribution associated
with these capability gaps and $\Delta H_g$ denotes the general-level edit.
General-level optimization is formulated as
\begin{equation}
H_g^{*}
=
\arg\max_{H_g \in S_g}
\mathbb{E}_{x \sim \mathcal{X}_{\mathcal{A}},\,
\tau \sim \mathcal{T}(H_g,H_t^{*}(x),x)}
\left[r_x(\Delta H_g, \tau)\right].
\label{eq:general_level_objective}
\end{equation}
The outer loop promotes improvements supported by the broader trace
history into $H_g$, where they persist across subsequent tasks.
The objectives above specify what each level seeks to improve; the
trace-grounded edits below provide the local improvement procedure.
Thus, although $M$ remains frozen, the effective capability of the
model--system pair can evolve through persistent changes to its
supporting system. Algorithm~\ref{alg:robofoundry} summarizes the overall
execution--evolution loop.
We detail diagnosis and improvement in
Section~\ref{sec:system_improvement}.

\begin{figure}[t]
    \centering
    \includegraphics[width=0.90\textwidth]{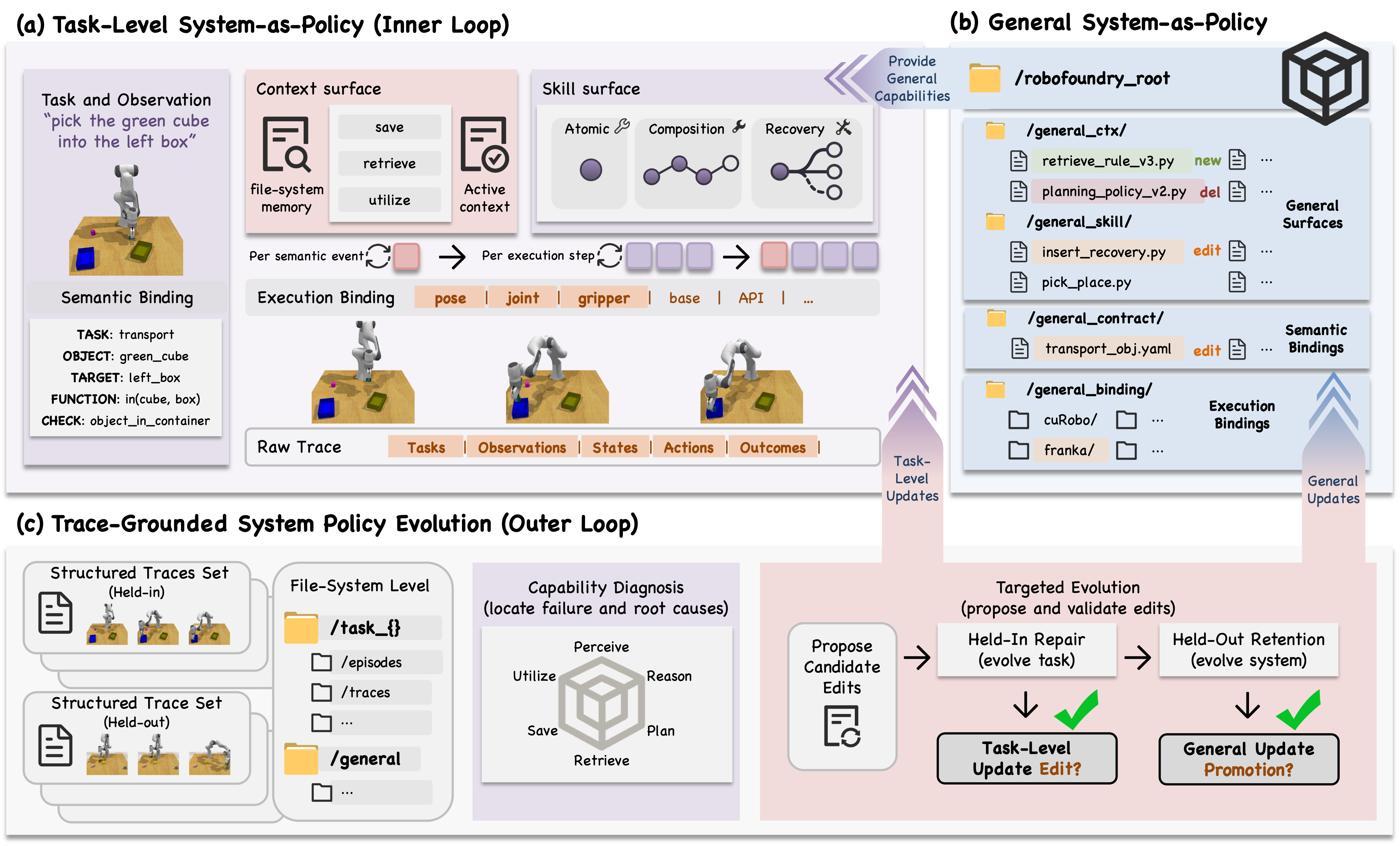}
    \caption{
    \textbf{Overview of RoboFoundry.} (a) The inner loop runs task-specific System-as-Policy through
    context/skill surfaces and semantic/execution bindings, producing embodiment-specific rollouts and traces.
    (b) The general system supplies reusable context and skill rules to task-level systems.
    (c) The outer loop diagnoses capability failures and validates
    targeted edits through held-in repair and held-out retention.
    Accepted edits update the task-specific system or promote the general system.}
    \label{fig:robo_model}
\end{figure}

\subsection{Inner-Loop System Execution}
\label{sec:inner_loop}

RoboFoundry uses the same task-execution loop in simulation and the
real world.
Given a task description, observations, and robot states,
cross-embodiment execution requires separating embodiment-invariant
task semantics from embodiment-specific kinematic and dynamic primitives.
We introduce a semantic binding layer
$H_t^s(M) = \texttt{BIND}_s(H_t(M)),$
which exposes task logic shared across embodiments.
Accordingly, $H_t^s(M)$ represents, reasons about, and decomposes the
task without directly handling robot-specific constraints.
An execution binding layer
$H_t^e(M) = \texttt{BIND}_e(H_t(M)),$
grounds the resulting semantic decisions in embodiment-specific execution.
This separation helps distinguish errors in task reasoning from failures
in execution.
The execution backend can use frozen VLA policies, functions generated
by coding agents, visuomotor APIs such as CuRobo, or other
foundation-model-supported robotic interfaces.

RoboFoundry exposes two task-level surfaces, \emph{context} and
\emph{skill}, for semantic-level optimization in $H_t^s(M)$.
(1) Context: The context system consists of persistent filesystem memory
and the active context presented to the model.
Rather than updating memory at every step, $H_t^s(M)$ monitors task
progress and updates context when an event warrants it.
We define the context operation space as
$S_t^{\mathrm{ctx}}
=
\{\texttt{SAVE},\texttt{RETRIEVE},\texttt{UTILIZE}\}$.
Given execution histories of observations, actions, and feedback,
\texttt{SAVE} selects task-relevant evidence for persistent storage.
What should be retained depends on task semantics: an occlusion task
may require object and spatial relations, whereas a counting task may
require occurrences of relevant actions.
\texttt{RETRIEVE} selects stored evidence for the next decision, and
\texttt{UTILIZE} incorporates it into the active context.
These operations are recorded in the execution trace as evidence for
subsequent system improvement.

(2) Skill: $\texttt{BIND}_s$ separates task logic from
embodiment-specific constraints.
We organize the skill surface hierarchically into atomic skills,
skill compositions, and recovery.
Atomic skills are indivisible executable units, such as \texttt{grasp}
or \texttt{move-to}, whose low-level implementations are initially
handcrafted and verified.
For a given task, atomic skills are composed into reusable procedures
that preserve its semantic logic; for example, \texttt{move-to},
\texttt{grasp}, and \texttt{place} can form a transfer procedure.
During the inner loop, RoboFoundry maintains a recovery tree with key
states as nodes and skill executions, including failed transitions,
as edges.
Upon failure, $H(M)$ queries the tree to identify a valid state from
which to replan, rather than repeating the failed action.
For example, if a cube falls during transfer, the system can use the
state where the cube rests on the support surface to plan another grasp.
Execution traces are therefore attributable to atomic execution,
composition, or recovery, providing evidence for skill improvement.

\subsection{Trace-Grounded Diagnosis and System Improvement}
\label{sec:system_improvement}

\paragraph{Task-level system improvement.}
Given an execution trace $\tau$, the task-level system policy $H_t(M)$
diagnoses the observed failure or inefficiency by attributing it to
one of six capabilities:
$\mathcal A =
\{
\texttt{perceive},
\texttt{reason},
\texttt{plan},\\
\texttt{save},
\texttt{retrieve},
\texttt{utilize}
\}$.
The first three characterize decision-making, while the latter three
characterize memory management.
$H_t(M)$ then localizes the capability gap and proposes an edit
$\Delta H_t$ to the corresponding context or skill surface.
The candidate is evaluated in a new inner-loop execution.
If it resolves the failure, we denote the successful edit by
$\Delta H_t^*=\Delta H_t$ and commit it as
$H_t^*\leftarrow H_t\oplus\Delta H_t^*$,
where $\oplus$ denotes applying an edit to the current system.
The resulting trace, including $\Delta H_t^*$ and its execution feedback,
is stored in $\mathcal T$ and indexed by capability $a\in\mathcal A$
for later general system improvement.

\paragraph{General system improvement.}
Task-specific corrections cannot guarantee performance on similar cases
when the underlying general capability remains unchanged.
RoboFoundry therefore evaluates whether a successful task-level edit
can be generalized into a reusable system-level rule.

For each successful repair, its stored trace $\tau_i$ records the task
$x_i$, observations and robot states before and after the repair,
pre-repair history, committed task-level edit $\Delta H_{t,i}^{*}$,
and execution feedback.
We select successful records of the same capability from other tasks
as the held-out trace set
$\mathcal T^{\mathrm{out}}_a(x)=
\{\tau_i\in\mathcal T:a_i=a,\ x_i\neq x\}$.

From the successful edit $\Delta H_t^*$ on task $x$ and its trace
$\tau_x$, we abstract a candidate general edit
$\Delta H_g\leftarrow
\operatorname{Abstract}(\Delta H_t^*,\tau_x;H_g)$.
For each held-out trace, let $\tau_i^{\mathrm{pre}}$ denote its recorded
task, observation, state, and pre-repair history.
We replay this information and use the candidate general system to
propose a task-level edit
$\Delta H_{t,i}^{'}\leftarrow
\operatorname{Propose}(H_g\oplus\Delta H_g,\tau_i^{\mathrm{pre}})$.
The system policy $H(M)$ then acts as a verifier.
Given the candidate $\Delta H_{t,i}^{'}$, the recorded successful edit
$\Delta H^*_{t,i}$, and its execution feedback, it scores whether the
candidate addresses the same capability gap.
Let $r_x(\Delta H_t,\tau_i)$ denote this score.
We measure capability-conditioned transfer as
\begin{equation}
\Delta_{\mathrm{cap}}(\Delta H_g)
=
\frac{1}{|\mathcal T^{\mathrm{out}}_a(x)|}
\sum_{\tau_i\in\mathcal T^{\mathrm{out}}_a(x)}
\left[
r_x(\Delta H_{t,i}^{'},\tau_i)-r_x(\Delta H_{t,i}^{*},\tau_i)
\right].
\label{eq:capability_transfer}
\end{equation}
We promote a candidate $\Delta H_g^{*}$ when
$\Delta_{\mathrm{cap}}(\Delta H_g^{*})\geq 0$ and commit it as
$H_g^{*}\leftarrow H_g\oplus\Delta H_g^{*}$.
Thus, task-level evaluation verifies whether an edit repairs the
observed failure, while capability-conditioned held-out evaluation
assesses whether the general edit addresses the same capability gap
across other tasks.

\begin{algorithm}[t]
\caption{RoboFoundry: system evolution and task execution}
\label{alg:robofoundry}
\footnotesize

\noindent
\begin{minipage}[t]{0.48\linewidth}
\vspace{0pt}
\textbf{Overall evolution}
\begin{algorithmic}[1]
\Require General system $H_g$, task-level system $H_t$,
         tasks $\mathcal X$, trace set $\mathcal T$
\State $\mathcal T\gets\emptyset$
\State \textbf{for} each task $x\in\mathcal X$ \textbf{do}
\State \hspace*{0.8em}\textbf{for} each rollout on $x$ \textbf{do}
\State \hspace*{1.6em}$\tau\gets
    \Call{RunTask}{H_g,H_t,x}$
\State \hspace*{1.6em}$(H_g,H_t,\mathcal T)\gets
    \Call{Revise}{H_g,H_t,\mathcal T,\tau}$
\State \Return $H_g, H_t ,\mathcal T$
\end{algorithmic}
\end{minipage}\hfill
\begin{minipage}[t]{0.48\linewidth}
\vspace{0pt}
\textbf{RunTask}$(H_g,H_t,x)$
\begin{algorithmic}[1]
\State $H_t^s\gets\texttt{BIND}_s(H_t)$;
       $H_t^e\gets\texttt{BIND}_e(H_t)$
\State Initialize state $z_0$, memory $\mathcal M_0$; $\tau\gets\emptyset$
\State \textbf{for} each step $t$ until completion \textbf{do}
\State \hspace*{0.8em}$c_t \gets
\textsc{Utilize}(\textsc{Retrieve}(x,z_t,\mathcal M_t),z_t)$
\State \hspace*{0.8em}$p_t\gets
    M(x,z_t,c_t;H_g,H_t^s)$
\State \hspace*{0.8em}$(z_{t+1},f_t)\gets
    \Call{Execute}{H_t^e(p_t)}$
\State \hspace*{0.8em}$\tau\gets\tau\mathbin{\|}
    (z_t,c_t,p_t,z_{t+1},f_t)$
\State \hspace*{0.8em}$\mathcal M_{t+1}\gets
    \Call{SAVE}{\mathcal M_t,\tau}$
    \textbf{ on event}
\State \Return $\tau$
\end{algorithmic}
\end{minipage}
\end{algorithm}

\section{Experiments}
\subsection{System-Level Evolution of Embodied Decision-Making}
\label{sec:embodiedbench}

\begin{wrapfigure}{R}{0.38\textwidth}
    \centering
    \includegraphics[width=\linewidth]
    {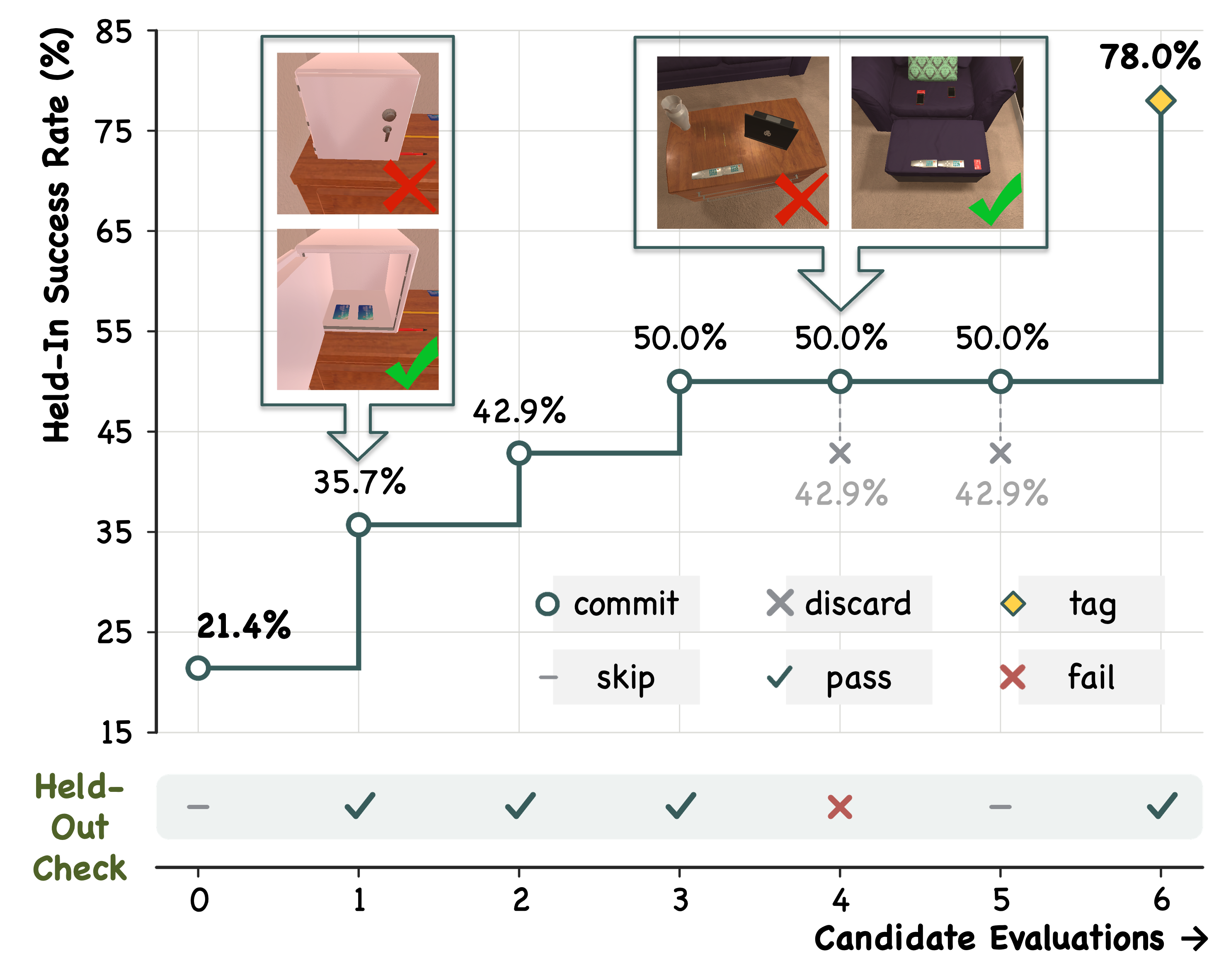}
    \caption{
    \textbf{Example of evolution trace on the EB-Habitat spatial subset.} Here, RoboFoundry (GPT-5.5) rises Held-in success rate from \SI{21.4}{\%} to \SI{78.0}{\%} over six candidate evaluations. The final version passes the held-out check and is promoted to the general system.
    }
    \label{fig:evolution_trace}
    \vspace{-0.5cm}
\end{wrapfigure}
EmbodiedBench~\citep{yang2025embodiedbench} evaluates embodied decision-making
across 1,128 tasks in four environments:
EB-ALFRED and EB-Habitat test high-level semantic planning, while
EB-Navigation and EB-Manipulation require low-level executable actions. We instantiate the raw backbone, \textit{RoboFoundry-Lite}, and full
\textit{RoboFoundry} across multiple frontier foundation models: GPT-6 Astra, GPT-5.5, Qwen3.7-Plus, Qwen3.8-27B, and GLM5.3-Flash.
Lite ablates general evolution and keeps task edits only.
All variants share the same task stream and interaction budget, evolve from
the first scored episode, and retain every evolution trajectory in the
reported score.

Table~\ref{tab:embodiedbench_main} shows consistent gains across all five
backbones, indicating that RoboFoundry is compatible with heterogeneous
foundation models and repairs backbone-specific bottlenecks rather than
converging to a fixed harness.
It improves GPT-5.5 by \SI{27.8}{\%}, and notably lifts Qwen3.7-Plus to
near parity with RoboFoundry (GPT-5.5), showing that performance is governed
by the evolved system rather than dominated by a stronger backbone.
Even the frontier-tier GPT-6 Astra benefits from RoboFoundry, with an
\SI{8.5}{\%} relative gain.
The \SI{11.2}{\%} relative gain of full RoboFoundry over RoboFoundry-Lite
on Qwen3.7-Plus further isolates general-scope evolution as the critical
ingredient. Subset results are detailed in
Tables~\ref{tab:ebalfred_main}--\ref{tab:ebmanipulation_main}. We further evaluate a 16K-token context budget in
Table~\ref{tab:embodiedbench_16k}. A larger budget itself does not consistently
improve task success (e.g., GLM5.3-Flash drops from 62.0\% at 4K to
61.3\% at 16K), while RoboFoundry still improves all backbones
at 16K by \SI{7.9}{\%}--\SI{13.1}{\%} relative to their baselines.

\begin{wrapfigure}{R}{0.38\textwidth}
    \centering
    \includegraphics[width=\linewidth]
    {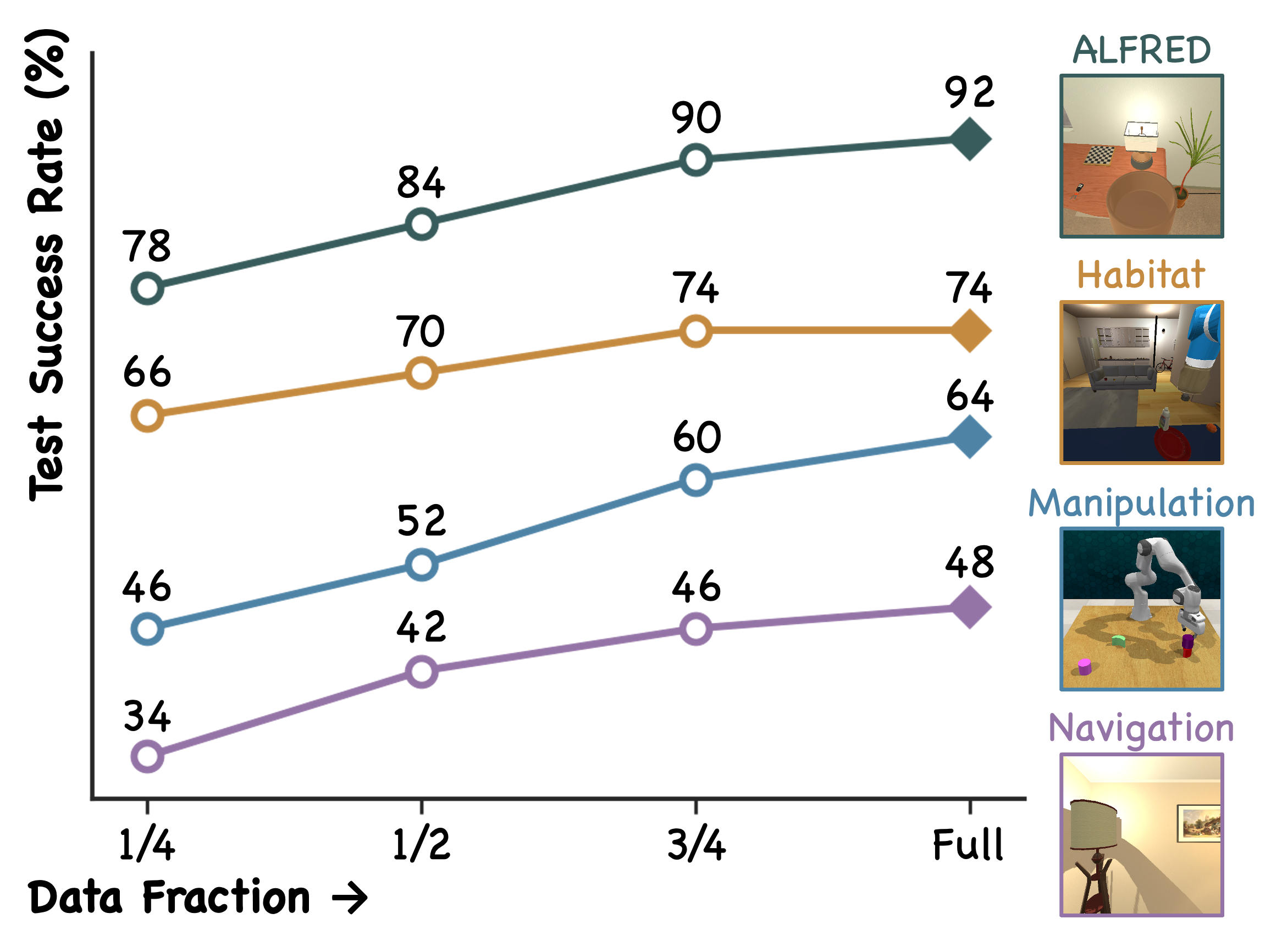}
    \caption{
    \textbf{Example of scaling with held-in data on EmbodiedBench.}
    RoboFoundry (GPT-5.5) is evaluated on 50 fixed held-out tasks
    per suite. Using three quarters of the held-in data already
    matches the full-data success rate of \SI{74.0}{\%} on EB-Habitat,
    highlighting data-efficient evolution.
    }
    \label{fig:data_fraction}
    \vspace{-0.5cm}
\end{wrapfigure}
Figure~\ref{fig:evolution_trace} further traces one evolution run on the EB-Habitat
spatial subset: three committed edits lift held-in success from
\SI{21.4}{\%} to \SI{50.0}{\%}, two candidates are discarded without changing
the system, and the last is tagged for promotion, reaching \SI{78.0}{\%}. We further investigate how RoboFoundry scales with the amount of held-in data
used for evolution, reserving 50 held-out tasks per suite for testing.
As shown in Figure~\ref{fig:data_fraction}, evolution is data-efficient:
one quarter of the held-in tasks already reaches \SI{78.0}{\%} on EB-ALFRED and
\SI{66.0}{\%} on EB-Habitat, and three quarters recover most of the final score,
with EB-Habitat saturating at \SI{74.0}{\%}.

\begin{table}[t]
    \centering
    \caption{
    \textbf{Overall comparison on EmbodiedBench.}
    Avg. is the arithmetic mean of success rates across the four suites.
    All results are reported in percentage (\%).
    Gains are reported in percentage points (+pp).
    }
    \label{tab:embodiedbench_main}

    \fontsize{7.5}{9.0}\selectfont
    \renewcommand{\arraystretch}{1.08}
    \setlength{\tabcolsep}{4.5pt}

    \providecommand{\ebgain}[1]{%
        \smash{\textsuperscript{\normalfont\fontsize{5}{5}\selectfont +#1}}%
    }

    \begin{tabularx}{\linewidth}{
        >{\raggedright\arraybackslash}X
        *{5}{>{\centering\arraybackslash}m{0.111\linewidth}}
    }
    \toprule
    \textbf{Method} &
    \textbf{Avg.} &
    \shortstack[c]{\textbf{EB-}\\\textbf{ALFRED}} &
    \shortstack[c]{\textbf{EB-}\\\textbf{Habitat}} &
    \shortstack[c]{\textbf{EB-}\\\textbf{Navigation}} &
    \shortstack[c]{\textbf{EB-}\\\textbf{Manipulation}} \\
    \midrule

    RoboFoundry (GPT-6 Astra) &
    \textbf{78.0}\ebgain{6.1} &
    \textbf{90.0}\ebgain{2.7} &
    86.7\ebgain{11.4} &
    \textbf{80.2}\ebgain{3.2} &
    \textbf{54.9}\ebgain{6.9} \\

    RoboFoundry (GPT-5.5) &
    72.7\ebgain{15.8} &
    84.0\ebgain{7.3} &
    \textbf{88.7}\ebgain{24.7} &
    72.0\ebgain{17.3} &
    45.9\ebgain{13.8} \\

    RoboFoundry (Qwen3.7-Plus) &
    70.3\ebgain{10.4} &
    81.3\ebgain{8.0} &
    80.3\ebgain{15.3} &
    71.9\ebgain{7.6} &
    47.7\ebgain{10.5} \\

    RoboFoundry (GLM5.3-Flash) &
    69.3\ebgain{7.4} &
    80.0\ebgain{8.3} &
    75.0\ebgain{14.3} &
    72.7\ebgain{1.7} &
    49.5\ebgain{5.1} \\

    RoboFoundry (Qwen3.8-27B) &
    66.5\ebgain{14.2} &
    78.0\ebgain{12.3} &
    77.0\ebgain{19.0} &
    68.3\ebgain{16.0} &
    42.8\ebgain{9.5} \\

    RoboFoundry-Lite (GPT-5.5) &
    65.5\ebgain{8.6} &
    78.0\ebgain{1.3} &
    74.3\ebgain{10.3} &
    69.7\ebgain{15.0} &
    39.9\ebgain{7.8} \\

    RoboFoundry-Lite (Qwen3.7-Plus) &
    63.2\ebgain{3.2} &
    75.0\ebgain{1.7} &
    72.0\ebgain{7.0} &
    68.0\ebgain{3.7} &
    37.6\ebgain{0.4} \\

    \midrule

    \mbox{GPT-6 Astra} &
    71.9 &
    87.3 &
    75.3 &
    77.0 &
    48.0 \\

    \mbox{GLM5.3-Flash} &
    62.0 &
    71.7 &
    60.7 &
    71.0 &
    44.4 \\

    \mbox{Qwen3.7-Plus} &
    60.0 &
    73.3 &
    65.0 &
    64.3 &
    37.2 \\

    \mbox{GPT-5.5} &
    56.9 &
    76.7 &
    64.0 &
    54.7 &
    32.1 \\

    \mbox{Qwen3.8-27B} &
    52.3 &
    65.7 &
    58.0 &
    52.3 &
    33.3 \\

    \mbox{Claude-3.5-Sonnet} &
    50.5 &
    64.0 &
    68.0 &
    44.7 &
    25.4 \\

    \mbox{GPT-4o} &
    50.4 &
    56.3 &
    59.0 &
    57.7 &
    28.5 \\

    \mbox{Claude-3.7-Sonnet} &
    49.9 &
    67.7 &
    58.7 &
    45.0 &
    28.3 \\

    \bottomrule
    \end{tabularx}
\end{table}

\subsection{Long-Horizon Memory through Context Evolution}
\label{sec:roboharness_robomemarena}

RoboMemArena~\citep{lei2026robomemarena} evaluates long-horizon memory across
26 tasks in four categories: \textbf{T}ransfer, \textbf{O}cclusion,
\textbf{C}ounting, and \textbf{S}equence.
We follow the official protocol and report task success rate (TSR) and
cumulative success rate (CSR).
RoboFoundry uses Qwen3.7-Plus as the backbone for the main comparison.
Evolution begins with the first scored episode, and all trajectories
used for updates count toward the evaluation budget.
Task state is reset between trials; only validated task-agnostic
context revisions persist.
Table~\ref{tab:robomemarena_full} shows that RoboFoundry achieves \SI{53.5}{\%} TSR and \SI{72.8}{\%} CSR, outperforming all baselines
across all four categories. Compared with PrediMem, RoboFoundry achieves its largest relative TSR
gain on Transfer, improving it by \SI{213.3}{\%}.

We further apply RoboFoundry to standalone
$\pi_{0.5}$~\citep{pmlr-v305-black25a} and PrediMem, which combines
Qwen3-VL-8B-Instruct for explicit memory management and planning with $\pi_{0.5}$ for execution.
Table~\ref{tab:robofoundry_memory_categories} shows improvements in both
TSR and CSR across all categories for both architectures, with a
\SI{200.0}{\%} relative TSR gain for PrediMem on Transfer.
These results support the applicability of RoboFoundry's context evolution across
different memory architectures.

\begin{table}[t]
    \centering
    \caption{
        \textbf{Long-horizon memory results on RoboMemArena.} RoboFoundry achieves the best TSR and CSR overall and across all four categories. Each entry reports TSR / CSR (\%).
    }
    \label{tab:robomemarena_full}

    \small
    \setlength{\tabcolsep}{3pt}
    \renewcommand{\arraystretch}{1.08}

    \begin{tabular}{lccccc}
        \toprule
        \textbf{Method}
        & \textbf{Overall}
        & \textbf{Transfer}
        & \textbf{Occlusion}
        & \textbf{Counting}
        & \textbf{Sequence} \\
        \midrule

        \rowcolor{gray!12}
        \textbf{RoboFoundry}
        & \textbf{53.5 / 72.8}
        & \textbf{70.5 / 78.1}
        & \textbf{38.3 / 57.6}
        & \textbf{55.2 / 82.5}
        & \textbf{75.2 / 92.1} \\

        PrediMem
        & 38.5 / 55.2
        & 22.5 / 45.2
        & 27.3 / 38.4
        & 45.7 / 69.3
        & 72.5 / 89.5 \\

        MemER
        & 27.3 / 49.1
        & 20.0 / 36.1
        & 16.4 / 33.2
        & 27.1 / 65.1
        & 65.0 / 79.1 \\

        HiF-VLA
        & 16.9 / 39.8
        & 17.5 / 38.9
        & 12.7 / 27.1
        & 8.6 / 45.9
        & 42.5 / 70.2 \\

        $\pi_{0.5}$
        & 21.5 / 38.7
        & 20.0 / 42.8
        & 12.7 / 17.2
        & 14.3 / 50.9
        & 60.0 / 71.6 \\

        MemoryVLA
        & 15.0 / 35.3
        & 15.0 / 37.2
        & 7.3 / 13.1
        & 14.3 / 55.1
        & 37.5 / 65.2 \\

        \bottomrule
    \end{tabular}
\end{table}

\begin{table}[t]
    \centering

    % ==================== Left: RoboMemArena ====================
    \begin{minipage}[t]{0.54\textwidth}
        \vspace{0pt}
        \centering

        \caption{
            \textbf{Context evolution across agent configurations.}
            RoboFoundry improves both standalone $\pi_{0.5}$ and PrediMem across all four categories. Each entry reports TSR/CSR (\%).
        }
        \label{tab:robofoundry_memory_categories}

        \scriptsize
        \setlength{\tabcolsep}{1.5pt}
        % Match the height of the adjacent LIBERO-PRO table.
        \renewcommand{\arraystretch}{1.7866}

        \begin{tabularx}{\linewidth}{
            >{\hsize=0.95\hsize\linewidth=\hsize\raggedright\arraybackslash}X
            >{\hsize=0.85\hsize\linewidth=\hsize\centering\arraybackslash}X
            >{\hsize=1.175\hsize\linewidth=\hsize\centering\arraybackslash}X
            >{\hsize=0.85\hsize\linewidth=\hsize\centering\arraybackslash}X
            >{\hsize=1.175\hsize\linewidth=\hsize\centering\arraybackslash}X
        }
            \toprule
            \multirow{2}{*}{\shortstack[l]{\textbf{Memory}\\\textbf{Category}}}
            & \multicolumn{2}{c}{$\boldsymbol{\pi}_{0.5}$}
            & \multicolumn{2}{c}{\textbf{PrediMem}} \\
            \cmidrule(lr){2-3}
            \cmidrule(lr){4-5}
            & \textbf{Base}
            & \textbf{+RoboFoundry}
            & \textbf{Base}
            & \textbf{+RoboFoundry} \\
            \midrule

            Transferring
            & 20.0 / 42.8
            & \textbf{62.5 / 71.2}
            & 22.5 / 45.2
            & \textbf{67.5 / 76.2} \\

            Occlusion
            & 12.7 / 17.2
            & \textbf{33.6 / 47.2}
            & 27.3 / 38.4
            & \textbf{36.4 / 56.4} \\

            Counting
            & 14.3 / 50.9
            & \textbf{58.8 / 79.6}
            & 45.7 / 69.3
            & \textbf{53.7 / 80.3} \\

            Sequence
            & 60.0 / 71.6
            & \textbf{70.0 / 90.8}
            & 72.5 / 89.5
            & \textbf{73.0 / 90.5} \\

            \midrule
            \textbf{Avg.}
            & 26.8 / 45.6
            & \textbf{56.2 / 72.2}
            & 42.0 / 60.6
            & \textbf{57.7 / 75.9} \\

            \bottomrule
        \end{tabularx}
    \end{minipage}
    \hfill
    % ==================== Right: LIBERO-PRO ====================
    \begin{minipage}[t]{0.44\textwidth}
        \vspace{0pt}
        \centering

        \caption{
            \textbf{Results on LIBERO-PRO under perturbations.}
            Each entry reports position/task success rate (\%).
            $^\dagger$ denotes privileged simulator object poses.
        }
        \label{tab:libero_pro_generalization}

        \scriptsize
        \setlength{\tabcolsep}{1.5pt}
        \renewcommand{\arraystretch}{1.16}

        \begin{tabularx}{\linewidth}{
            >{\hsize=1.3\hsize\linewidth=\hsize\raggedright\arraybackslash}X
            *{3}{>{\hsize=0.9\hsize\linewidth=\hsize\centering\arraybackslash}X}
        }
            \toprule
            \textbf{Method}
            & \textbf{Object}
            & \textbf{Goal}
            & \textbf{Spatial} \\
            \midrule

            \shortstack[l]{
                OpenVLA
            }
            & 0.0 / 0.0
            & 0.0 / 0.0
            & 0.0 / 0.0 \\

            \shortstack[l]{
                $\pi_{0}$
            }
            & 0.0 / 0.0
            & 0.0 / 0.0
            & 0.0 / 0.0 \\

            \shortstack[l]{
                $\pi_{0.5}$
            }
            & 17.0 / 1.0
            & 38.0 / 0.0
            & 20.0 / 1.0 \\

            \shortstack[l]{
                CaP-Agent0
            }
            & 21.8 / 18.2
            & 25.6 / 16.8
            & 11.8 / 14.0 \\
            \shortstack[l]{ASPIRE}
            & \textbf{98.0 / 95.0}
            & \textbf{81.0 / 45.0}
            & \textbf{51.0 / 60.0} \\
            \shortstack[l]{Harness VLA \\(Codex)}
            & \textbf{81.0 / 69.0}
            & \textbf{94.0 / 91.0} 
            & \textbf{75.0 / 66.0} \\
            \shortstack[l]{Harness VLA \\(CC)}
            & \textbf{94.0 / 80.0}
            & \textbf{88.0 / 90.0}
            & \textbf{87.0 / 87.5} \\
            \midrule
            % \textbf{RoboFoundry}
            % & \textbf{66.0 / 68.0}
            % & \textbf{48.0 / 46.0}
            % & \textbf{42.0 / 41.5} \\
            \shortstack[l]{\textbf{RoboFoundry}}
            & \textbf{96.0 / 98.0}
            & \textbf{88.0 / 86.0}
            & \textbf{92.0 / 91.5} \\

            \shortstack[l]{RoboFoundry$^\dagger$}
            & 99.0 / 100.0
            & 96.0 / 100.0
            & 98.0 / 99.2 \\

            \bottomrule
        \end{tabularx}
    \end{minipage}
\end{table}

\subsection{System Evolution under Distribution Shift}
\label{sec:libero_pro}

We use LIBERO-PRO~\citep{zhou2025libero} to test whether persistent system evolution improves
robustness to changes in object layouts and task specifications. We compare against CaP-Agent0~\citep{fu2026capx},
ASPIRE~\citep{lu2026aspire}, and both Harness VLA variants
(CodeX and CC)~\citep{zhang2026harness}. Following the CaP-Agent0 protocol, we evaluate 30 tasks
from the Object, Goal, and Spatial suites under
\emph{Position} and \emph{Task} perturbations, using the same perception
and control primitives, multi-turn interaction setting, and 50 trials
per task and perturbation.
Evolution begins with the first scored rollout, and all rollouts used
for updates count toward the evaluation budget.

Table~\ref{tab:libero_pro_generalization} shows that RoboFoundry
achieves the highest average success rate across the six settings
among methods without privileged object poses
(\SI{91.9}{\%} vs. \SI{87.8}{\%} for Harness VLA (CC)).
It also outperforms CaP-Agent0 with up to a
\SI{679.7}{\%} relative gain on Spatial position perturbations, supporting the benefit of persistent system
evolution beyond within-episode program repair.
We report task-wise results in
Tables~\ref{tab:libero-pro-object-taskwise}--\ref{tab:libero-pro-spatial-taskwise}.
\begin{figure}[t]
    \centering
    \includegraphics[width=\textwidth]{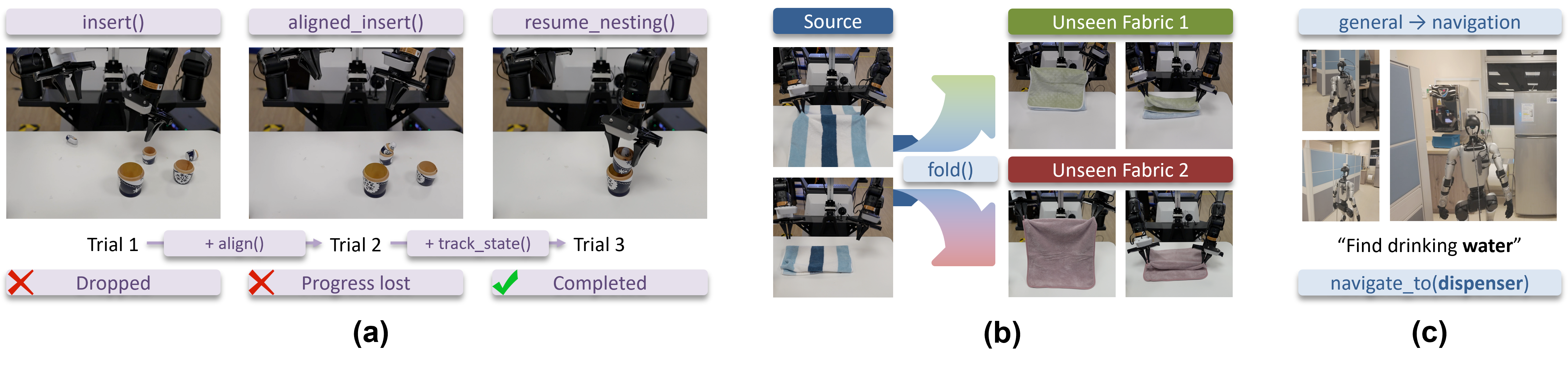}
    \caption{
    \textbf{Real-world evolution and transfer of RoboFoundry.}
    (a) Failures motivate alignment and progress-tracking edits,
    enabling nesting-doll completion in Trial~3.
    (b) Towel-folding skills evolved on the blue towel transfer
    to unseen green and pink fabrics.
    (c) The general system grounds a semantic navigation goal
    in executable actions on a Unitree G1 humanoid.
    }
    \label{fig:real-doll-towel}
\end{figure}

To assess how much perception limits the evolved system, we additionally
provide RoboFoundry$^\dagger$ with privileged simulator object poses.
Success improves across all six settings, reaching \textbf{100\%}
on both Object and Goal task perturbations.
These gains suggest that perception remains a bottleneck and that
RoboFoundry can translate more accurate state information into
more reliable task execution.

\subsection{Real-World Evaluation}
\label{sec:real_world}
\begin{wrapfigure}{r}{0.38\textwidth}
    \centering
    \includegraphics[width=\linewidth]
    {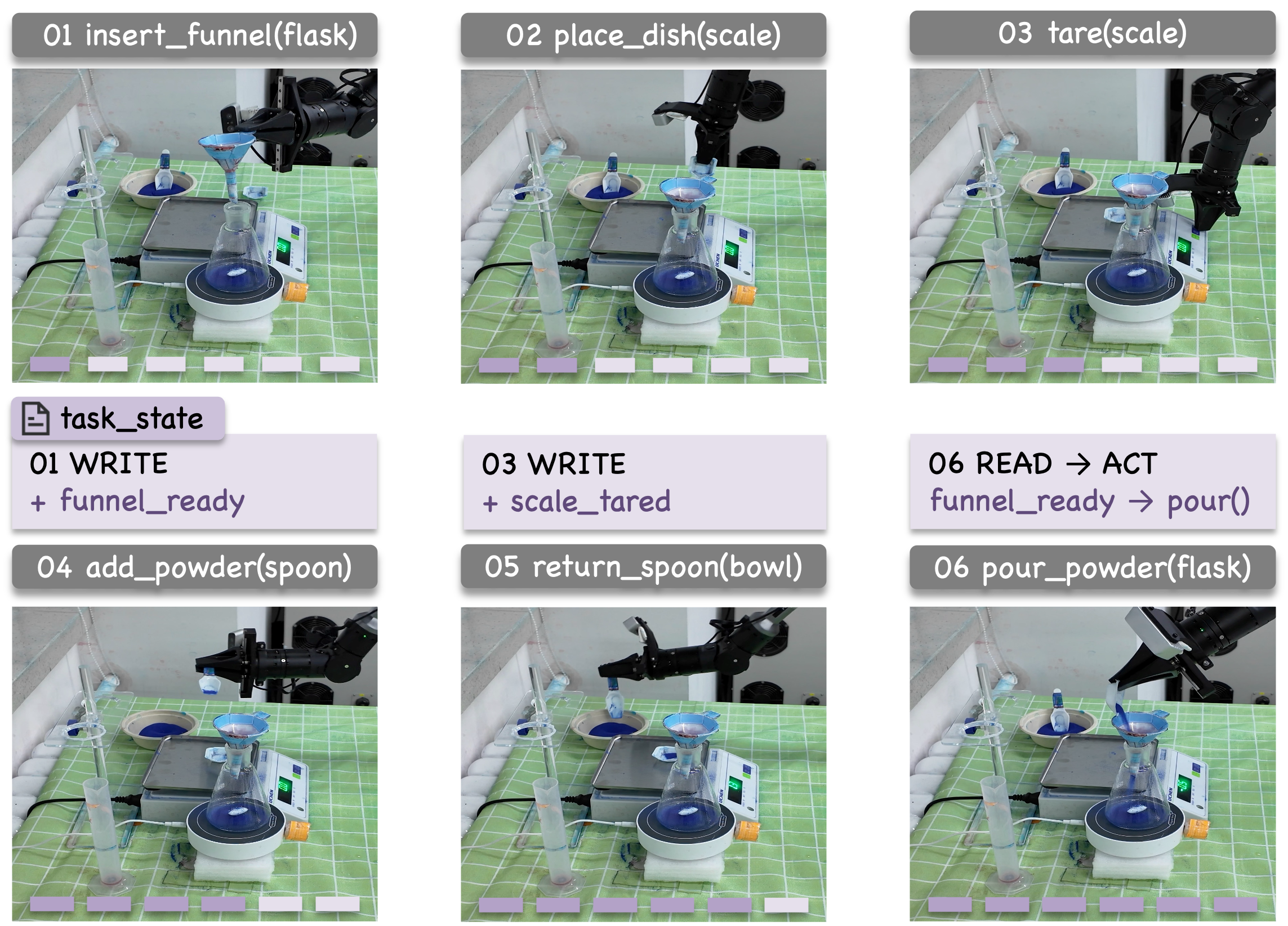}
    \caption{
    \textbf{Long-horizon chemistry manipulation with RoboFoundry.}
    A task-state record connects six dependent subtasks: funnel readiness
    and scale taring are written during execution, and the final pouring
    step reads the earlier funnel-ready state to guide skill execution.
    }
    \label{fig:real-nav-chem}
    \vspace{-0.5cm}
\end{wrapfigure}
We evaluate whether the general RoboFoundry system transfers to physical
robots and continues improving after deployment.
Throughout the real-world experiments, RoboFoundry takes GPT-5.5 as the supporting foundation model and $\pi_{0.5}$ as its manipulation backend,
following~\citep{lei2026robomemarena}.
For a fair comparison, RoboFoundry and the standalone baseline $\pi_{0.5}$ share
the same fine-tuned checkpoint on corresponding tasks.
RoboFoundry starts without task-specific system adaptation.
We report the success rate of 10 trials per task, including all trials used
for online evolution, with system updates applied only between trials.

We first evaluate online evolution on an AgileX COBOT MAGIC platform. As shown in Figure~\ref{fig:real-doll-towel}(a), nesting-doll manipulation requires both small-to-large ordering and
precise insertion. Early failures drive complementary skill and context revisions:
alignment checks improve insertion, while progress tracking enables
execution to resume from the last completed subtask. Finally, RoboFoundry achieves \SI{70}{\%} success, compared with
\SI{30}{\%} for standalone $\pi_{0.5}$.

Moreover, Figure~\ref{fig:real-doll-towel}(b) demonstrates zero-shot transfer
of the evolved folding skill: RoboFoundry evolves only on the blue
towel and transfers directly to held-out green and pink towels
with different appearance, geometry, and material properties. Without further adaptation, RoboFoundry achieves over \SI{80}{\%} success, compared with below \SI{20}{\%} for $\pi_{0.5}$, which needs further fine-tuning.
Figure~\ref{fig:real-doll-towel}(c) extends RoboFoundry beyond manipulation
to semantic navigation on a Unitree G1 humanoid.
Through the shared semantic interface, RoboFoundry maps ``Find drinking
water'' to a dispenser identified in RGB observations, while
embodiment-specific bindings translate this target into navigation actions.
As shown in Figure~\ref{fig:real-nav-chem}, we also evaluate a long-horizon chemistry experiment with six dependent
subtasks. RoboFoundry successfully preserves subtask state through its context surface and
coordinates execution through reusable skill compositions and recovery
structures. Crucially, completed steps become persistent prerequisites for later
actions: the final pouring step retrieves the funnel-ready state
recorded several subtasks earlier. On the other hand, standalone $\pi_{0.5}$ struggles to complete the full procedure.

\section{Related Work}

\paragraph{Foundation Models as Embodied Agents.}
LLMs and MLLMs increasingly guide embodied decision-making.
Early work connects language and multimodal observations to robot
actions through vision-language-action models, multimodal prompts,
and feedback-guided planning
~\citep{brohan2023can,pmlr-v202-jiang23b,pmlr-v205-huang23c}.
Code-as-policy methods further enable models to compose perception
and control primitives into executable programs
~\citep{fu2026capx,zhang2026playful}.
The same backbone can behave differently depending on its surrounding
interaction harness~\citep{liu2026guava,zhang2026harness},
verification~\citep{kwok2026llm}, and embodied decision
interface~\citep{li2024embodied}.
RoboFoundry formalizes this dependence as \emph{System-as-Policy}:
the executable policy is jointly determined by the foundation model
and its supporting system. It makes this system an explicit target of
evolution, turning embodied experience into persistent changes to
future behavior.

\paragraph{Self-Evolving LLM/MLLM Agents.}
A parallel line improves agents through reflective
self-correction~\citep{madaan2023self,shinn2023reflexion},
persistent memory~\citep{packer2023memgpt,zhao2024expel},
and automated search over prompts, workflows, or
code~\citep{khattab2023dspy,hu2025automated,zhang2026darwin}.
RoboFoundry extends system-level adaptation to cross-embodied
execution. Its shared semantic interface separates task-level
reasoning from robot-specific execution, allowing system revisions
to be evaluated and reused across embodiments.

\paragraph{Self-Improving Embodied Agents.}
Existing work studies skill acquisition
~\citep{wang2023voyager,lu2026aspire,wang2026skillmemo},
memory-guided execution and policy orchestration
~\citep{huang2026roboharness,zhang2026harness,lei2026robomemarena},
program repair with execution feedback~\citep{fu2026capx},
and closed-loop policy improvement
~\citep{li2026roboclaw,xiao2026enpire}.
RoboFoundry treats context and skills as complementary intervention
surfaces of the executable system policy. It evaluates task-level
repairs on related held-out cases before promoting them into general
system capabilities that support subsequent tasks.

\section{Conclusion}
\label{sec:conclusion}
We present RoboFoundry, an embodied agentic framework for \emph{System-as-Policy Evolution}. Acting as a bridge between general-purpose models and embodied scenarios, RoboFoundry improves the embodied agent lifecycle, from decision-making and memory management to skill execution.
It uses execution traces to improve task performance and promotes validated improvements into general system capabilities that benefit future tasks. RoboFoundry demonstrates strong autonomy and evolvability across multiple benchmarks and diverse real-world tasks. Looking forward, we aim to build an embodied flywheel that connects simulation and real-world experience, transfers knowledge from diverse non-embodied data to embodied tasks, and enables system evolution and model learning to reinforce each other.

\bibliographystyle{plainnat}
\bibliography{paper}

\clearpage
\appendix
\section{Appendix}

\newtcblisting{tracebox}{
    listing only,
    enhanced,
    breakable,
    colback=gray!2,
    colframe=black!20,
    boxrule=0.3pt,
    arc=1pt,
    left=3pt,
    right=3pt,
    top=1.5pt,
    bottom=1.5pt,
    before skip=3pt,
    after skip=3pt,
    listing options={
        basicstyle=\ttfamily\scriptsize,
        columns=fullflexible,
        keepspaces=true,
        showstringspaces=false,
        breaklines=true,
        aboveskip=0pt,
        belowskip=0pt
    }
}

\newtcolorbox{gitdiffbox}{
    enhanced,
    breakable,
    colback=gray!2,
    colframe=black!20,
    boxrule=0.3pt,
    arc=1pt,
    left=4pt,
    right=4pt,
    top=1.5pt,
    bottom=1.5pt,
    before skip=3pt,
    after skip=3pt,
    fontupper=\ttfamily\scriptsize,
    parskip=0pt
}

\newcommand{\gadd}[1]{%
    {\color{green!45!black}\ttfamily\scriptsize
    \detokenize{+ #1}}\par
}

\newcommand{\gdel}[1]{%
    {\color{red!65!black}\ttfamily\scriptsize
    \detokenize{- #1}}\par
}

\newcommand{\gmod}[1]{%
    {\color{orange!75!black}\ttfamily\scriptsize
    \detokenize{M #1}}\par
}

\newcommand{\gctx}[1]{%
    {\color{black!80}\ttfamily\scriptsize
    \detokenize{  #1}}\par
}

\newcommand{\ghead}[1]{%
    {\color{black}\bfseries\ttfamily\scriptsize #1}\par
}

\subsection{EmbodiedBench Evolution Case Studies}
\label{sec:supp_embodiedbench_evolution}
\begin{figure}[H]
    \centering
    \includegraphics[width=\linewidth]
    {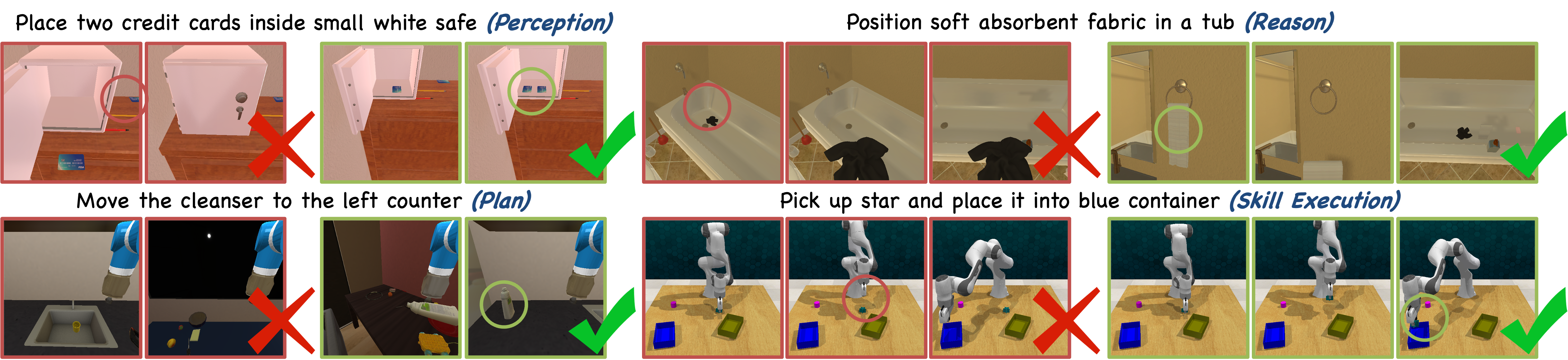}
    \caption{
    \textbf{Representative EmbodiedBench cases (GPT-5.5) before and after
    RoboFoundry evolution.}
    Examples illustrate repairs to perception, reasoning, planning, and
    low-level skill execution.
    }
    \label{fig:embodiedbench-cases}
\end{figure}

This section provides several examples of how RoboFoundry evolves on EmbodiedBench. Figure~\ref{fig:embodiedbench-cases} illustrates four representative
failures repaired through changes to perception, reasoning, planning,
and skill execution. We compare the initial execution (GPT-5.5) with the final candidate promoted by RoboFoundry's evolution. Specifically, we expose three levels of change:
(1) the persistent \emph{general} system, (2) its \emph{task-specific} realization during execution, and (3) the corresponding changes in the filesystem. To ensure a fair comparison, we use the same episode and instruction for the initial and final candidates.

\subsubsection{Example 1: Feedback-Grounded Recovery in EB-Habitat}
\label{sec:supp_habitat_evolution}

Consider the instruction
\texttt{"Relocate every cleanser from sink onto the sofa."}
The initial system permits relatively long action sequences.
The accepted revision instead introduces observation checkpoints after
navigation and treats environment feedback as stronger evidence than
visual appearance for interaction feasibility:

\begin{gitdiffbox}
\gdel{Each plan should include no more than 20 actions.}
\gadd{Prefer one action; at most two when both are currently supported.}
\gadd{After Navigation, observe before Pick/Place/Open/Close.}
\gadd{Trust environment feedback over visual appearance for proximity.}
\gadd{After a not-near Pick failure, navigate before retrying.}
\end{gitdiffbox}

This is a shared cross-task modification: it does not encode the
cleanser, sink, sofa, or other entities from the proposal episode.
During execution, the shared rule is instantiated using task-conditioned
feedback. The persistent system policy stores the reusable recovery rule, while
the runtime instantiation binds it to the current failed action, object,
and environment feedback:

\begin{gitdiffbox}
\gdel{Pick(cleanser) -> not near -> repeated retries -> failure}
\gadd{Pick(cleanser) -> not near}
\gadd{Navigate(table 2) -> Pick(cleanser) -> Navigate(sofa) -> Place(sofa)}
\gadd{task success}
\end{gitdiffbox}

\subsubsection{Example 2: Skill Refinement in EB-Manipulation}
\label{sec:supp_manip_evolution}

We further examine
\texttt{"Please put the red star into the shape sorter."}
The initial rollout executes 15 valid actions without completing the
task, whereas the rollout under the accepted revision succeeds at the
sixth action. The revision introduces an explicit skill-level insertion
strategy:

\begin{gitdiffbox}
\gadd{For shape-sorter tasks, insert the object into the matching opening.}
\gadd{Keep holding while aligning and descending into the opening.}
\gadd{Release only when the object is inside or just below the opening.}
\end{gitdiffbox}

The corresponding task-conditioned action sequence changes from an early
release to a grasp-preserving descent.
Actions follow
\texttt{[x,y,z,rx,ry,rz,gripper]}, where the final dimension controls
the gripper:

\begin{gitdiffbox}
\gdel{[49,60,48,0,60,45,0] // above opening, closed}
\gdel{[49,60,40,0,60,45,1] // early release}
\gadd{[49,60,48,0,60,35,0] // above opening, closed}
\gadd{[49,60,39,0,60,35,0] // descend while holding}
\gadd{[49,60,38,0,60,35,1] // low-position release}
\end{gitdiffbox}

\subsubsection{Example 3: Support-Surface Grounding in EB-ALFRED}
\label{sec:supp_alfred_evolution}

The instruction
\texttt{"Place the newspaper next to the left laptop"}
requires distinguishing a semantic spatial reference from a physically
valid support surface.
The laptop defines the target spatial region, while the sofa provides
the executable support surface. The accepted revision explicitly
separates these roles in the runtime plan:

\begin{gitdiffbox}
\gctx{find Newspaper}
\gctx{pick up Newspaper}
\gctx{find Laptop}
\gadd{find Sofa}
\gctx{put down}
\end{gitdiffbox}

\subsubsection{Example 4: Reachable-Floor Reasoning in EB-Navigation}
\label{sec:supp_navigation_evolution}

For
\texttt{"navigate to the Toaster and be as close as possible"},
the initial planner primarily reasons about the target location and
nearby obstacles. The planner under the accepted revision also uses
shorter movement segments, allowing new observations and collision
feedback to affect subsequent actions.
The final target distance decreases from \SI{3.9}{m} to \SI{0.9}{m}.
The accepted revision adds an explicit rule for reasoning over reachable
floor:

\begin{gitdiffbox}
\gdel{Reason based on the target object's location and nearby obstacles.}
\gadd{Reason about the reachable floor path, not only the target image location.}
\gadd{For targets on furniture, approach the free floor beside the support surface.}
\end{gitdiffbox}

\subsubsection{Persistent Filesystem Evolution}
\label{sec:supp_filesystem_evolution}

RoboFoundry does not treat evolution as transient model output.
A candidate becomes an \emph{accepted revision} only after evaluation;
accepted revisions are then committed to the agent workspace.
For EmbodiedBench, the execution and outer-loop systems
are organized as follows:

\begin{tracebox}
workspace/
+-- overlay/embodiedbench/
| +-- evaluator/config/
| | +-- system_prompts.py
| | +-- habitat_examples.json
| | +-- eb_navigation_examples.json
| | +-- eb_manipulation_examples.json
| | +-- eb_alfred_examples.json
| +-- planner/
| +-- vlm_planner.py
| +-- nav_planner.py
| +-- manip_planner.py
+-- .robofoundry/
+-- AGENTS.md
+-- READ_POLICY.md
+-- bootstrap/{snapshot.json, summary.md}
+-- experience/{parent_summary.json, parent/...}
+-- <benchmark>_feedback/{summary.md, failures.json}
\end{tracebox}

\paragraph{Execution surface.}
\texttt{overlay/embodiedbench/} contains components directly used
during agent execution.
Evolution modifies persistent rules, demonstrations, and execution feedback. These modifications constitute the persistent system policy reused by subsequent executions. Representative file modifications are:

\begin{gitdiffbox}
\ghead{ALFRED}
\gmod{evaluator/config/system_prompts.py}
\gmod{planner/planner_utils.py}
\gmod{planner/vlm_planner.py}

\ghead{Habitat}
\gmod{evaluator/config/system_prompts.py}
\gmod{evaluator/config/habitat_examples.json}
\gmod{planner/vlm_planner.py}

\ghead{Navigation}
\gmod{evaluator/config/system_prompts.py}
\gmod{evaluator/config/eb_navigation_examples.json}
\gmod{planner/nav_planner.py}

\ghead{Manipulation}
\gmod{evaluator/config/system_prompts.py}
\gmod{evaluator/config/eb_manipulation_examples.json}
\gmod{planner/manip_planner.py}
\end{gitdiffbox}

\paragraph{Outer-loop evolution evidence.}
The \texttt{.robofoundry/} branch is the filesystem realization of the
Evolution Ledger $\mathcal L$.
It stores information used by the outer evolution process, including
bootstrap state, parent--candidate information, evaluation summaries,
and benchmark-specific failures. Observations, actions, feedback, and model outputs remain
associated with evaluation traces:

\begin{tracebox}
evaluation/
+-- results/episode_*_res.json
+-- episode_*_step_*.json
+-- traces/episode_*/
    +-- step_*/
        +-- prompt.txt
        +-- model_output.txt
\end{tracebox}

These execution traces provide the grounded evidence from which
candidate system revisions are diagnosed and evaluated.
Taken together, the filesystem separates two roles: runtime execution
produces grounded traces, while accepted revisions persist reusable
decision rules and execution constraints back into the system policy.

\subsection{RoboMemArena Evolution Case Studies}
\label{sec:supp_robomem_evolution}

\begin{figure}[!t]
    \centering
    \includegraphics[width=0.95\textwidth]{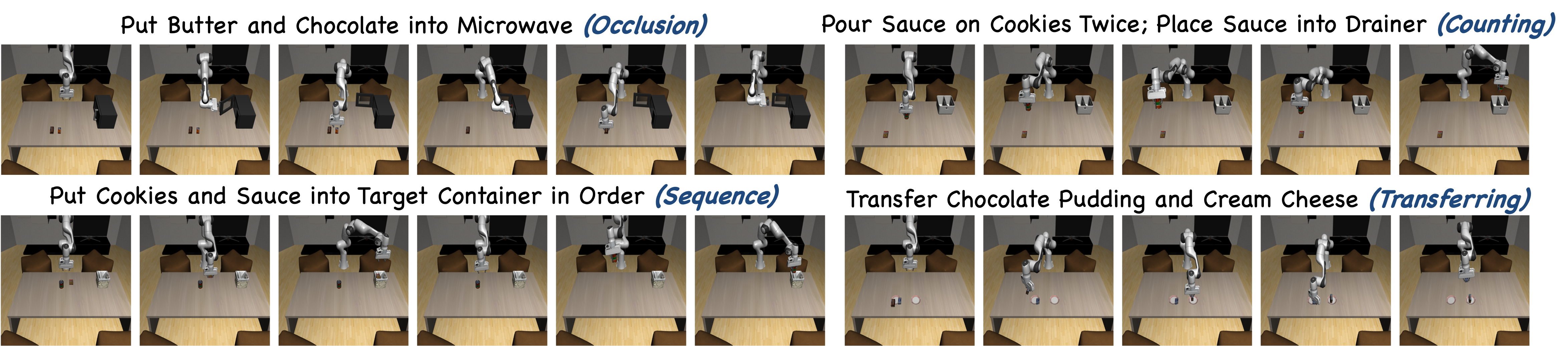}
    \caption{RoboFoundry rollouts across four memory categories in
    RoboMemArena: Transfer, Occlusion, Counting, and Sequence.}
    \label{fig:robomem-rollouts}
\end{figure}

We provide several examples of how RoboFoundry changes the
context-management system on RoboMemArena. We expose how an
accepted system revision changes
(1) persistent context or memory rules,
(2) the task-conditioned input, and
(3) the corresponding files in the agent workspace.
We select representative tasks from four RoboMemArena categories:
Transfer, Occlusion, Counting, and Sequence.
Figure~\ref{fig:robomem-rollouts} shows the corresponding rollouts
and illustrates how the evolved context supports task execution.

\subsubsection{Example 1: Task Progress as Active Context in Transfer}
\label{sec:supp_robomem_transfer}

We first examine a Transferring task that requires moving both
\emph{butter} and \emph{cream cheese} from one plate to another.
The initial formatter ignores runtime progress and always returns the
original task prompt, whereas the revised formatter explicitly selects
the next unfinished stage:

\begin{gitdiffbox}
\gdel{del context}
\gdel{return str(base_prompt)}
\gadd{completed_count = _completed_stage_count(context)}
\gadd{stage = _next_stage_name(context)}
\gadd{overall = "overall task: <base task>"}
\gadd{return <next stage> + <overall task> + "do not undo completed steps"}
\end{gitdiffbox}

The persistent rule is task-agnostic: object names appear only in its
runtime instantiation from the current execution state.
For the selected episode:

\begin{tracebox}
Runtime state:
  01_Place_Butter_Plate2        = true
  02_Place_Cream_Cheese_Plate2 = false
\end{tracebox}

This changes the actual input from a static task description to
an explicit active subgoal:

\begin{gitdiffbox}
\gdel{butter cream}
\gadd{place cream cheese plate 2.}
\gadd{overall task: butter cream.}
\gadd{do not undo completed steps}
\end{gitdiffbox}

The evolved context therefore preserves the active subgoal, global task,
and completed progress simultaneously.
The selected episode progresses from completing only the butter stage
to completing both stages; across the corresponding 10-episode search
evaluation, full-task success increases from \SI{40}{\%} to \SI{70}{\%}.

\subsubsection{Example 2: External Visual Memory under Occlusion}
\label{sec:supp_robomem_occlusion}

The Occlusion example requires manipulating cookies with a microwave
and later placing popcorn into it.
The evolved memory system introduces explicit rules for combining the
current observation, sparse visual history, and attempted trials:

\begin{gitdiffbox}
\gadd{Ground decisions in the latest frame first.}
\gadd{Use historical keyframes for persistent objects, receptacles, and progress.}
\gadd{Keep frames that capture persistent state changes or useful locations.}
\gadd{Attempted commands are not evidence of physical success.}
\gadd{If the current image supports completion, advance to the next primitive.}
\end{gitdiffbox}

The memory selector additionally provides sparse fallback anchors when
no selected keyframes are available and summarizes recent attempted
commands into the active context.
For one selected invocation, the retrieved visual context is:

\begin{tracebox}
memory_indices_before = [0, 576, 672]
recent_window         = [711, ..., 715]
previous_primitive    = open microwave
place cookies         = attempted 31x (t=560--710)
\end{tracebox}

The corresponding contextual guidance emphasizes that the retrieved
historical frames are reminders rather than authoritative evidence:

\begin{gitdiffbox}
\gadd{Use sparse historical frames as reminders of persistent scene state.}
\gadd{Compare them with the current frame before deciding to continue or move on.}
\gadd{Attempted VLA prompts are not evidence that an action succeeded.}
\end{gitdiffbox}

The initially selected episode repeatedly returns to
\texttt{pick cookies}, whereas the evolved execution later transitions
to \texttt{pick popcorn} and \texttt{place popcorn}.
Recorded stage completion changes from \SI{50}{\%} to \SI{75}{\%}.
This example illustrates the evolution of both which historical
observations are retrieved and how retrieved evidence is
presented and interpreted.

\subsubsection{Example 3: Counting as Stage-Conditioned Context}
\label{sec:supp_robomem_counting}

The Counting task requires two pouring operations followed by a final
placement.
The initial formatter always returns the same base prompt.
The revised formatter instead converts ordered stage state into an
explicit next-step instruction:

\begin{gitdiffbox}
\gadd{completed_count = number of completed stages}
\gadd{unfinished_count = number of remaining stages}
\gadd{append "<N> earlier steps already complete"}
\gadd{append "Next, <next unfinished stage>"}
\gadd{if one stage remains: finish it and avoid redoing completed steps}
\end{gitdiffbox}

Across the search evaluation, full-task success increases
from \SI{60}{\%} to \SI{80}{\%}. After both pouring stages have been recorded as complete, the runtime
prompt changes to:

\begin{gitdiffbox}
\gdel{pour tomato sauce into bowl drainer}
\gadd{Task: pour tomato sauce into bowl drainer.}
\gadd{2 earlier steps are already complete.}
\gadd{Next, place bowl drainer.}
\gadd{Complete this remaining step now and avoid redoing completed steps.}
\end{gitdiffbox}

\subsubsection{Example 4: Local Progress Improvement in Sequence}
\label{sec:supp_robomem_sequence}

RoboFoundry changes the context system and selects key episodes. The evolved format exposes the active stage and adds stage-dependent execution constraints:

\begin{gitdiffbox}
\gadd{For place/put/insert/move: the named item should end at the destination.}
\gadd{For open/close: change only the named fixture for this step.}
\gadd{Do not redo completed steps.}
\gadd{After all stages complete: preserve the final arrangement.}
\end{gitdiffbox}

One runtime prompt becomes:

\begin{tracebox}
Current step: place cookies microwave.
Full task: pour tomato cookies microwave.
The named item should end at the named destination.
Do not redo completed steps.
\end{tracebox}

After all listed stages are marked complete:

\begin{tracebox}
Full task: pour tomato cookies microwave.
All listed steps are complete;
hold the final arrangement and avoid disturbing objects.
\end{tracebox}

\subsubsection{Persistent Filesystem Changes}
\label{sec:supp_robomem_filesystem}

The behavior changes above correspond to persistent modifications in
the execution workspace rather than transient natural-language
suggestions:

\begin{tracebox}
candidates/<candidate>/workspace/overlay/
+-- evaluation_benchmark/
+-- scripts/
| +-- robofoundry_policy.py
+-- async_vlm_reference/
+-- memory_policy.py
+-- eval_async_vlm_vla.py
\end{tracebox}

The stage-conditioned Transferring, Counting, and Sequence cases modify
the shared context-management implementation, while the Occlusion case
additionally changes visual-memory selection, retrieval, and active-context
construction:

\begin{gitdiffbox}
\ghead{Transferring / Counting / Sequence}
\gmod{evaluation_benchmark/scripts/robofoundry_policy.py}

\ghead{Occlusion}
\gmod{evaluation_benchmark/async_vlm_reference/memory_policy.py}
\gmod{evaluation_benchmark/async_vlm_reference/eval_async_vlm_vla.py}
\end{gitdiffbox}

The persistent changes cover context formatting, ordered-stage
processing, stage-conditioned context construction, replanning cadence,
historical-frame selection, and memory-context formatting.

\subsubsection{Evolution Evidence and Execution Traces}
\label{sec:supp_robomem_evidence}

Outer-loop evolution evidence is stored separately from the
execution-facing system.
The \texttt{.robofoundry/} branch is the filesystem realization:

\begin{tracebox}
workspace/
+-- .robofoundry/
    +-- AGENTS.md
    +-- READ_POLICY.md
    +-- bootstrap/{snapshot.json, summary.md}
    +-- experience/
    |   +-- parent_summary.json
    |   +-- parent/{manifest.json, evaluation/, validation/}
    +-- <adapter>_feedback/
        +-- summary.md
        +-- failures.json
        +-- image_evidence/*.png
\end{tracebox}

Task-conditioned evidence remains attached to each evaluation episode:

\begin{tracebox}
candidates/<candidate>/evaluation/<adapter>/run/taskN/epM/
+-- episode_result.json
+-- stage_events.jsonl
+-- vla_input_trace.jsonl       # runtime prompt / context
+-- vla_prompt_trace.jsonl      # VLA prompt
+-- sync_vlm_trace.jsonl        # VLM decision / memory indices
+-- vla_inputs/
+-- vlm_inputs/
\end{tracebox}

This organization separates three roles: execution traces record
episode-level state, actions, and failures; evolution evidence supports
diagnosis and candidate evaluation; and accepted revisions persist
reusable context-selection, retrieval, and utilization rules back into
the system policy.

These cases expose complementary forms of context management.
Transferring and Counting convert recorded progress into an explicit
active subgoal; Occlusion changes the retrieval and utilization of
external visual memory; and Sequence shows that better structured
active context can improve intermediate progress without resolving the
complete task.
Together, these examples show that RoboFoundry evolves not only the
content of a prompt, but also the mechanisms that determine what state
is retained, what evidence is retrieved, and how it is presented to the
embodied model.

\subsection{LIBERO-PRO Evolution Case Studies}
\label{sec:supp_libero_pro_evolution}

We further inspect how RoboFoundry revises the executable system on
LIBERO-PRO.
In contrast to the context-centric changes observed in RoboMemArena,
the main revisions occur at the interface between high-level task
reasoning and physical execution, i.e., code-as-policy.
We distinguish two adaptation scopes.
Persistent shared evolution modifies prompt templates, API
semantics, reusable execution skills, and logic shared across tasks.
In contrast, trial-local recovery occurs within an individual
trial, where generated task code is revised using the current
observation, simulator state, execution feedback, and previous failures.

\subsubsection{Persistent Evolution}
\label{sec:supp_libero_general}

The selected persistent revision modifies four files in the shared
workspace. These changes jointly modify task generation, API semantics, reusable
execution skills, and environment launch:
\begin{tracebox}
candidates/<id>/workspace/
+-- env_configs/libero/
| +-- franka_libero_robofoundry.yaml
+-- robofoundry/integrations/franka/
| +-- libero_reduced.py
| +-- libero_harness.py
+-- robofoundry/utils/
+-- launch_utils.py
\end{tracebox}

\paragraph{Task-generation prompt.}
The evolved template changes how ordinary object transfer should be
implemented. The persistent rule specifies \emph{how} a transfer should be executed
without encoding object-specific cases:

\begin{gitdiffbox}
\gdel{Prefer CuRobo grasp and transport planning with small fallbacks.}
\gadd{Prefer pick_and_place_sim_object(...) for ordinary object transfer.}
\gadd{Pass the desired final object-center XYZ from the live target pose.}
\gadd{Use a modest positive Z clearance for receptacles or support surfaces.}
\gadd{Keep TCP offsets and bounded grasp retries inside the helper.}
\gadd{Use lower-level CuRobo only when scene geometry requires it.}
\end{gitdiffbox}

\paragraph{Coordinate semantics.}
The revised execution interface consistently separates the desired
object-center target from the TCP target used by the controller.
This keeps hand-to-object offsets inside the embodiment-specific
execution interface rather than repeatedly re-estimating them in
generated task code:

\begin{tracebox}
object_position = get_sim_object_pose(object_name)[0]
tcp_position    = get_ee_pose()[0]
tcp_object_offset = tcp_position - object_position

tcp_target =
    desired_object_position + tcp_object_offset
\end{tracebox}

\paragraph{Reusable helpers.}
The evolved system exposes three higher-level operations:

\begin{tracebox}
grasp_sim_object(...)
place_held_sim_object(...)
pick_and_place_sim_object(...)
\end{tracebox}

The grasp skill reads the current object pose and performs bounded
physical grasp attempts.
The placement skill converts the requested object-center position into
a TCP target and performs hover, descent, release, and retreat.
The composite skill connects these operations.
These skills execute physical motions rather than directly setting
simulator object state.

\subsubsection{Example 1: From Long Manipulation Code to Composite Skills}
\label{sec:supp_libero_object}

For task
\texttt{"Pick the cream cheese and place it in the basket."}, the initial
generated solution explicitly constructs segmentation, grasp generation,
trajectory planning, lift validation, transport, descent, and release.
Under the persistent shared revision, the generated task code instead
binds live object poses to the reusable manipulation skills:

\begin{gitdiffbox}
\gdel{segment cream cheese and generate grasp poses}
\gdel{plan_grasp_trajectory(...) and execute trajectory}
\gdel{manually compute transport offset, descend, and release}
\gadd{cream_key = match cream cheese from live object poses}
\gadd{basket_pos = get_sim_object_pose(basket_key)[0]}
\gadd{target = basket_pos + [0, 0, 0.12]}
\gadd{pick_and_place_sim_object(cream_key, target, ...)}
\end{gitdiffbox}

The persistent abstraction alone does not immediately complete the
task. Furthermore, feedback-driven rewrites introduce a trial-local recovery.
The composite manipulation skill is therefore persisted in the shared
system, whereas the push--regrasp recovery remains local to this trial.
This exposes two distinct adaptation time scales rather than a single
monolithic code update:

\begin{tracebox}
read current cream-cheese and basket poses
        |
        v
construct a tangential push direction
        |
        v
push object and verify XY displacement
        |
        v
grasp_sim_object(...)
        |
        v
place_held_sim_object(...)
\end{tracebox}

\subsubsection{Example 2: Execution Abstraction without Spatial Grounding Success}
\label{sec:supp_libero_spatial}

For task instruction
\texttt{"Pick the akita black bowl next to the ramekin and place it on the plate."},
both the initial and revised systems use simulator poses to identify the
bowl nearest to the ramekin.
The main persistent change is therefore the abstraction of the
subsequent manipulation:

\begin{gitdiffbox}
\gctx{select bowl nearest to ramekin using current XY distance}
\gdel{segment bowl}
\gdel{generate grasp poses}
\gdel{plan_grasp_trajectory(...)}
\gdel{manually lift, transport, descend, and release}
\gadd{target_pos = plate_pos + vertical clearance}
\gadd{pick_and_place_sim_object(bowl_key, target_pos, ...)}
\end{gitdiffbox}

\subsubsection{Filesystem Organization}
\label{sec:supp_libero_filesystem}

The LIBERO-PRO workspace separates the persistent shared system,
outer-loop evolution evidence, and trial-local generated task code.

\paragraph{Persistent executable system.}
The archived persistent revision modifies the task-generation interface,
API semantics, reusable skills, and launch logic:

\begin{gitdiffbox}
\gmod{env_configs/libero/franka_libero_robofoundry.yaml}
\gmod{robofoundry/integrations/franka/libero_reduced.py}
\gmod{robofoundry/integrations/franka/libero_harness.py}
\gmod{robofoundry/utils/launch_utils.py}
\end{gitdiffbox}

\paragraph{Outer evolution evidence.}
The \texttt{.robofoundry/} branch is the filesystem realization of the
Evolution Ledger $\mathcal L$ and stores evidence used by the outer evolution process. The \textit{curobo\_debug} files are motion-planning debug artifacts:

\begin{tracebox}
workspace/
+-- .robofoundry/
|   +-- AGENTS.md
|   +-- bootstrap/{snapshot.json, summary.md}
|   +-- experience/
|   |   +-- parent_summary.json
|   |   +-- parent/{manifest.json, evaluation/, validation/}
|   +-- libero_pro_feedback/{report.md, summary.json}
+-- curobo_debug/
    +-- *.npz
    +-- *.obj
\end{tracebox}

\paragraph{Task-specific execution.}
Generated task code and its feedback-driven revision history remain
inside the corresponding evaluation trial:

\begin{tracebox}
candidates/<id>/evaluation/libero_pro/cells/<suite>/task_01/
+-- result.json
+-- stdout.txt
+-- stderr.txt
+-- config.yaml
+-- output/robofoundry/run/
    +-- trial_index.json
    +-- trial_01_.../
        +-- trial_manifest.json
        +-- code.py
        +-- summary.txt
        +-- motion_debug.json
        +-- codegen_trace/
            +-- initial/{prompt.txt, final/output.txt}
            +-- regen_*/{prompt.txt, final/output.txt}
\end{tracebox}

The filesystem thus makes the two adaptation scopes explicit:
persistent shared evolution stores reusable execution abstractions in
the shared workspace, whereas trial-local recovery remains inside the
corresponding trial and revises generated task code from execution
feedback.

\begin{table}[t]
\centering
\caption{
\textbf{Main comparison on the EB-ALFRED high-level benchmark.}
Avg. denotes the arithmetic mean over Base, Common Sense,
Complex Instruction, Visual, Spatial, and Long Horizon.
All results are reported in percentage (\%).
}
\label{tab:ebalfred_main}

\fontsize{7.5}{9.0}\selectfont
\renewcommand{\arraystretch}{1.08}
\setlength{\tabcolsep}{3.5pt}

\begin{tabularx}{\linewidth}{
  >{\raggedright\arraybackslash}m{0.30\linewidth}
  *{3}{>{\hsize=0.95\hsize\linewidth=\hsize\centering\arraybackslash}X}
  >{\hsize=1.30\hsize\linewidth=\hsize\centering\arraybackslash}X
  *{3}{>{\hsize=0.95\hsize\linewidth=\hsize\centering\arraybackslash}X}
}
\toprule
\textbf{Method} &
\textbf{Avg.} &
\textbf{Base} &
\shortstack[c]{\textbf{Common}\\\textbf{Sense}} &
\shortstack[c]{\textbf{Complex}\\\textbf{Instruction}} &
\textbf{Visual} &
\textbf{Spatial} &
\shortstack[c]{\textbf{Long}\\\textbf{Horizon}} \\
\midrule

RoboFoundry (GPT-6 Astra) &
\textbf{90.0} & \textbf{92.0} & \textbf{92.0} &
\textbf{92.0} & \textbf{86.0} & \textbf{94.0} & 84.0 \\

RoboFoundry (GPT-5.5) &
84.0 & 90.0 & 84.0 & 84.0 & 74.0 & 80.0 & \textbf{92.0} \\

RoboFoundry (Qwen3.7-Plus) &
81.3 & 88.0 & 90.0 & 72.0 & 74.0 & 82.0 & 82.0 \\

RoboFoundry (GLM5.3-Flash) &
80.0 & 82.0 & 76.0 & 82.0 & 80.0 & 74.0 & 86.0 \\

RoboFoundry (Qwen3.8-27B) &
78.0 & 82.0 & 84.0 & 82.0 & 72.0 & 68.0 & 80.0 \\

RoboFoundry-Lite (GPT-5.5) &
78.0 & 82.0 & 80.0 & 76.0 & 70.0 & 72.0 & 88.0 \\

RoboFoundry-Lite (Qwen3.7-Plus) &
75.0 & 76.0 & 80.0 & 82.0 & 74.0 & 62.0 & 76.0 \\

\midrule

\mbox{GPT-6 Astra} &
87.3 & \textbf{92.0} & \textbf{92.0} &
\textbf{92.0} & 80.0 & 88.0 & 80.0 \\

\mbox{GPT-5.5} &
76.7 & 86.0 & 82.0 & 82.0 & 70.0 & 72.0 & 68.0 \\

\mbox{Qwen3.7-Plus} &
73.3 & 82.0 & 78.0 & 66.0 & 78.0 & 56.0 & 80.0 \\

\mbox{GLM5.3-Flash} &
71.7 & 76.0 & 76.0 & 76.0 & 70.0 & 62.0 & 70.0 \\

\mbox{Claude-3.7-Sonnet} &
67.7 & 68.0 & 68.0 & 70.0 & 68.0 & 62.0 & 70.0 \\

\mbox{Qwen3.8-27B} &
65.7 & 72.0 & 76.0 & 60.0 & 62.0 & 68.0 & 56.0 \\

\mbox{Claude-3.5-Sonnet} &
64.0 & 72.0 & 66.0 & 76.0 & 60.0 & 58.0 & 52.0 \\

\mbox{GPT-4o} &
56.3 & 64.0 & 54.0 & 68.0 & 46.0 & 52.0 & 54.0 \\

\bottomrule
\end{tabularx}
\end{table}

\begin{table}[t]
\centering
\caption{
\textbf{Main comparison on the EB-Habitat high-level benchmark.}
Avg. denotes the arithmetic mean over Base, Common Sense,
Complex Instruction, Visual, Spatial Relationship, and Long Horizon.
All results are reported in percentage (\%).
}
\label{tab:ebhabitat_main}

\fontsize{7.5}{9.0}\selectfont
\renewcommand{\arraystretch}{1.08}
\setlength{\tabcolsep}{3.5pt}

\begin{tabularx}{\linewidth}{
  >{\raggedright\arraybackslash}m{0.30\linewidth}
  *{3}{>{\hsize=0.95\hsize\linewidth=\hsize\centering\arraybackslash}X}
  >{\hsize=1.30\hsize\linewidth=\hsize\centering\arraybackslash}X
  *{3}{>{\hsize=0.95\hsize\linewidth=\hsize\centering\arraybackslash}X}
}
\toprule
\textbf{Method} &
\textbf{Avg.} &
\textbf{Base} &
\shortstack[c]{\textbf{Common}\\\textbf{Sense}} &
\shortstack[c]{\textbf{Complex}\\\textbf{Instruction}} &
\textbf{Visual} &
\textbf{Spatial} &
\shortstack[c]{\textbf{Long}\\\textbf{Horizon}} \\
\midrule

RoboFoundry (GPT-5.5) &
\textbf{88.7} & \textbf{100.0} & 78.0 & 80.0 &
86.0 & \textbf{100.0} & \textbf{88.0} \\

RoboFoundry (GPT-6 Astra) &
86.7 & \textbf{100.0} & \textbf{100.0} & \textbf{100.0} &
\textbf{100.0} & 42.0 & 78.0 \\

RoboFoundry (Qwen3.7-Plus) &
80.3 & \textbf{100.0} & 80.0 & 92.0 &
88.0 & 48.0 & 74.0 \\

RoboFoundry (Qwen3.8-27B) &
77.0 & \textbf{100.0} & 58.0 & 80.0 &
80.0 & 94.0 & 50.0 \\

RoboFoundry (GLM5.3-Flash) &
75.0 & \textbf{100.0} & 58.0 & 66.0 &
80.0 & 94.0 & 52.0 \\

RoboFoundry-Lite (GPT-5.5) &
74.3 & \textbf{100.0} & 66.0 & 74.0 &
80.0 & 44.0 & 82.0 \\

RoboFoundry-Lite (Qwen3.7-Plus) &
72.0 & 98.0 & 66.0 & 82.0 &
80.0 & 42.0 & 64.0 \\

\midrule

\mbox{GPT-6 Astra} &
75.3 & 88.0 & 84.0 & 86.0 & 82.0 & 38.0 & 74.0 \\

\mbox{Claude-3.5-Sonnet} &
68.0 & 96.0 & 68.0 & 78.0 & 70.0 & 38.0 & 58.0 \\

\mbox{Qwen3.7-Plus} &
65.0 & 92.0 & 58.0 & 76.0 & 74.0 & 34.0 & 56.0 \\

\mbox{GPT-5.5} &
64.0 & 94.0 & 54.0 & 62.0 & 70.0 & 32.0 & 72.0 \\

\mbox{GLM5.3-Flash} &
60.7 & 94.0 & 54.0 & 54.0 & 64.0 & 34.0 & 64.0 \\

\mbox{GPT-4o} &
59.0 & 86.0 & 44.0 & 56.0 & 68.0 & 36.0 & 64.0 \\

\mbox{Claude-3.7-Sonnet} &
58.7 & 90.0 & 58.0 & 58.0 & 62.0 & 38.0 & 46.0 \\

\mbox{Qwen3.8-27B} &
58.0 & 96.0 & 46.0 & 48.0 & 60.0 & 36.0 & 62.0 \\

\bottomrule
\end{tabularx}
\end{table}

\begin{table}[t]
\centering
\caption{
\textbf{Main comparison on the EB-Navigation benchmark.}
Avg. denotes the arithmetic mean over Base, Common Sense,
Complex Instruction, Visual, and Long Horizon.
All results are reported in percentage (\%).
}
\label{tab:ebnavigation_main}

\fontsize{7.5}{9.0}\selectfont
\renewcommand{\arraystretch}{1.08}
\setlength{\tabcolsep}{3.5pt}

\begin{tabularx}{\linewidth}{
  >{\raggedright\arraybackslash}m{0.30\linewidth}
  *{3}{>{\hsize=0.94\hsize\linewidth=\hsize\centering\arraybackslash}X}
  >{\hsize=1.30\hsize\linewidth=\hsize\centering\arraybackslash}X
  *{2}{>{\hsize=0.94\hsize\linewidth=\hsize\centering\arraybackslash}X}
}
\toprule
\textbf{Method} &
\textbf{Avg.} &
\textbf{Base} &
\shortstack[c]{\textbf{Common}\\\textbf{Sense}} &
\shortstack[c]{\textbf{Complex}\\\textbf{Instruction}} &
\textbf{Visual} &
\shortstack[c]{\textbf{Long}\\\textbf{Horizon}} \\
\midrule

RoboFoundry (GPT-6 Astra) &
\textbf{80.3} & \textbf{90.0} & \textbf{85.0} &
\textbf{86.7} & 73.3 & \textbf{66.7} \\

RoboFoundry (GLM5.3-Flash) &
72.7 & 80.0 & 71.7 & 78.3 & 73.3 & 60.0 \\

RoboFoundry (GPT-5.5) &
72.0 & 80.0 & 83.3 & 80.0 & 76.7 & 40.0 \\

RoboFoundry (Qwen3.7-Plus) &
72.0 & 78.3 & 83.3 & 65.0 & 66.7 & \textbf{66.7} \\

RoboFoundry-Lite (GPT-5.5) &
69.7 & 83.3 & 81.7 & 76.7 & \textbf{80.0} & 26.7 \\

RoboFoundry (Qwen3.8-27B) &
68.3 & 76.7 & \textbf{85.0} & 73.3 & 63.3 & 43.3 \\

RoboFoundry-Lite (Qwen3.7-Plus) &
68.0 & 78.3 & 76.7 & 68.3 & 63.3 & 53.3 \\

\midrule

\mbox{GPT-6 Astra} &
77.0 & 86.7 & 83.3 & 83.3 & 71.7 & 60.0 \\

\mbox{GLM5.3-Flash} &
71.0 & 80.0 & 73.3 & 80.0 & 66.7 & 55.0 \\

\mbox{Qwen3.7-Plus} &
64.3 & 75.0 & 73.3 & 63.3 & 60.0 & 50.0 \\

\mbox{GPT-4o} &
57.7 & 55.0 & 60.0 & 58.3 & 60.0 & 55.0 \\

\mbox{GPT-5.5} &
54.7 & 68.3 & 66.7 & 61.7 & 65.0 & 11.7 \\

\mbox{Qwen3.8-27B} &
52.3 & 41.7 & 66.7 & 60.0 & 53.3 & 40.0 \\

\mbox{Claude-3.7-Sonnet} &
45.0 & 50.0 & 61.7 & 50.0 & 36.7 & 26.7 \\

\mbox{Claude-3.5-Sonnet} &
44.7 & 66.7 & 51.7 & 41.7 & 36.7 & 26.7 \\

\bottomrule
\end{tabularx}
\end{table}

\begin{table}[t]
\centering
\caption{
\textbf{Main comparison on the EB-Manipulation benchmark.}
Avg. denotes the success rate over all 228 tasks.
All results are reported in percentage (\%).
}
\label{tab:ebmanipulation_main}

\fontsize{7.5}{9.0}\selectfont
\renewcommand{\arraystretch}{1.08}
\setlength{\tabcolsep}{3.5pt}

\begin{tabularx}{\linewidth}{
  >{\raggedright\arraybackslash}m{0.30\linewidth}
  *{3}{>{\hsize=0.94\hsize\linewidth=\hsize\centering\arraybackslash}X}
  >{\hsize=1.30\hsize\linewidth=\hsize\centering\arraybackslash}X
  *{2}{>{\hsize=0.94\hsize\linewidth=\hsize\centering\arraybackslash}X}
}
\toprule
\textbf{Method} &
\textbf{Avg.} &
\textbf{Base} &
\shortstack[c]{\textbf{Common}\\\textbf{Sense}} &
\shortstack[c]{\textbf{Complex}\\\textbf{Instruction}} &
\textbf{Visual} &
\textbf{Spatial} \\
\midrule

RoboFoundry (GPT-6 Astra) &
\textbf{54.8} & \textbf{58.3} & \textbf{62.5} &
\textbf{56.3} & 55.6 & 41.7 \\

RoboFoundry (GLM5.3-Flash) &
49.6 & 56.3 & 45.8 &
\textbf{56.3} & 47.2 & 41.7 \\

RoboFoundry (Qwen3.7-Plus) &
47.8 & 43.8 & 54.2 & 52.1 & 44.4 & 43.8 \\

RoboFoundry (GPT-5.5) &
46.1 & 54.2 & 47.9 & 43.8 & 41.7 & 41.7 \\

RoboFoundry (Qwen3.8-27B) &
42.5 & 35.4 & 41.7 & 43.8 & 47.2 & \textbf{45.8} \\

RoboFoundry-Lite (GPT-5.5) &
39.5 & 43.8 & 37.5 & 31.3 & 47.2 & 39.6 \\

RoboFoundry-Lite (Qwen3.7-Plus) &
37.3 & 33.3 & 41.7 & 31.3 & 44.4 & 37.5 \\

\midrule

\mbox{GPT-6 Astra} &
48.2 & 56.3 & 58.3 & 47.9 & 44.4 & 33.3 \\

\mbox{GLM5.3-Flash} &
43.0 & 41.7 & 41.7 & 33.3 & \textbf{72.2} & 33.3 \\

\mbox{Qwen3.7-Plus} &
36.0 & 35.4 & 35.4 & 27.1 & 61.1 & 27.1 \\

\mbox{Qwen3.8-27B} &
32.0 & 31.3 & 31.3 & 22.9 & 58.3 & 22.9 \\

\mbox{GPT-5.5} &
30.7 & 29.2 & 29.2 & 22.9 & 58.3 & 20.8 \\

\mbox{GPT-4o} &
28.9 & 39.6 & 29.2 & 29.2 & 19.4 & 25.0 \\

\mbox{Claude-3.7-Sonnet} &
28.5 & 31.3 & 20.8 & 43.8 & 25.0 & 20.8 \\

\mbox{Claude-3.5-Sonnet} &
25.4 & 37.5 & 16.7 & 29.2 & 19.4 & 22.9 \\

\bottomrule
\end{tabularx}
\end{table}

\begin{table}[t]
\centering
\caption{
\textbf{Overall comparison on EmbodiedBench (16K).}
Avg. is the arithmetic mean of success rates across the four suites.
All results are reported in percentage (\%).
Gains are reported in percentage points (+pp).
}
\label{tab:embodiedbench_16k}
\fontsize{7.5}{9.0}\selectfont
\renewcommand{\arraystretch}{1.08}
\setlength{\tabcolsep}{4.5pt}

\newcommand{\ebgain}[1]{%
  \smash{\textsuperscript{\normalfont\fontsize{5}{5}\selectfont +#1}}%
}

\begin{tabularx}{\linewidth}{
    >{\raggedright\arraybackslash}X
    *{5}{>{\centering\arraybackslash}m{0.111\linewidth}}
}
\toprule
\textbf{Method} &
\textbf{Avg.} &
\shortstack[c]{\textbf{EB-}\\\textbf{ALFRED}} &
\shortstack[c]{\textbf{EB-}\\\textbf{Habitat}} &
\shortstack[c]{\textbf{EB-}\\\textbf{Navigation}} &
\shortstack[c]{\textbf{EB-}\\\textbf{Manipulation}} \\
\midrule

RoboFoundry (GPT-5.5) &
\textbf{71.7}\ebgain{8.3} &
82.3\ebgain{3.3} &
\textbf{81.3}\ebgain{12.3} &
75.7\ebgain{8.4} &
47.3\ebgain{9.0} \\

RoboFoundry (Qwen3.7-Plus) &
70.4\ebgain{5.8} &
\textbf{82.5}\ebgain{2.8} &
79.3\ebgain{5.6} &
71.8\ebgain{1.0} &
\textbf{47.8}\ebgain{13.7} \\

RoboFoundry (Qwen3.8 27B) &
69.4\ebgain{5.1} &
79.7\ebgain{3.4} &
79.3\ebgain{8.0} &
75.0\ebgain{6.0} &
43.7\ebgain{3.0} \\

RoboFoundry (GLM5.3-Flash) &
68.8\ebgain{7.6} &
80.0\ebgain{5.7} &
72.0\ebgain{5.3} &
\textbf{77.7}\ebgain{8.7} &
45.5\ebgain{10.5} \\

\midrule

\mbox{Qwen3.7-Plus} &
64.6 &
79.7 &
73.7 &
70.8 &
34.1 \\

\mbox{Qwen3.8-27B} &
64.3 &
76.3 &
71.3 &
69.0 &
40.7 \\

\mbox{GPT-5.5} &
63.4 &
79.0 &
69.0 &
67.3 &
38.3 \\

\mbox{GLM5.3-Flash} &
61.3 &
74.3 &
66.7 &
69.0 &
35.0 \\

\bottomrule
\end{tabularx}
\end{table}

\begin{table}[htbp]
  \centering
  \caption{Task-wise LIBERO-PRO performance on the \texttt{libero-object}
  benchmark. Each entry reports position/task success rate in percentage (\%). $\dagger$
  denotes the use of privileged simulator object poses. Results for OpenVLA, $\pi_0$, $\pi_{0.5}$, and CaP-Agent0 are taken
directly from the CaP-Agent0 report~\citep{fu2026capx}.
  }
  \label{tab:libero-pro-object-taskwise}
  \setlength{\tabcolsep}{3.0pt}
  \renewcommand{\arraystretch}{1.08}
  \resizebox{\textwidth}{!}{%
  \begin{tabular}{l*{12}{c}}
    \toprule
    \textbf{Task (Symbolic Form)}
      & \multicolumn{2}{c}{\textbf{OpenVLA}}
      & \multicolumn{2}{c}{$\boldsymbol{\pi_0}$}
      & \multicolumn{2}{c}{$\boldsymbol{\pi_{0.5}}$}
      & \multicolumn{2}{c}{\textbf{CaP-Agent0}}
      & \multicolumn{2}{c}{\textbf{RoboFoundry}}
      & \multicolumn{2}{c}{\textbf{RoboFoundry}$^{\boldsymbol{\dagger}}$} \\
    \cmidrule(lr){2-3}\cmidrule(lr){4-5}\cmidrule(lr){6-7}
    \cmidrule(lr){8-9}\cmidrule(lr){10-11}\cmidrule(lr){12-13}
      & Pos & Task & Pos & Task & Pos & Task & Pos & Task
      & Pos & Task & Pos & Task \\
    \midrule

    $\operatorname{Place}(\texttt{alphabet\_soup},\texttt{basket})$
      & 0.0 & 0.0 & 0.0 & 0.0 & 0.0 & 0.0 & 2.0 & 4.0 & 98.0 & 99.0 & 100.0 & 100.0 \\
    $\operatorname{Place}(\texttt{bbq\_sauce},\texttt{basket})$
      & 0.0 & 0.0 & 0.0 & 0.0 & 100.0 & 2.0 & 12.0 & 42.0 & 97.0 & 93.0 & 100.0 & 100.0 \\
    $\operatorname{Place}(\texttt{butter},\texttt{basket})$
      & 0.0 & 0.0 & 0.0 & 0.0 & 54.0 & 0.0 & 26.0 & 18.0 & 97.0 & 99.0 & 100.0 & 100.0 \\
    $\operatorname{Place}(\texttt{chocolate\_pudding},\texttt{basket})$
      & 0.0 & 0.0 & 0.0 & 0.0 & 0.0 & 2.0 & 18.0 & 48.0 & 88.0 & 93.0 & 100.0 & 100.0 \\
    $\operatorname{Place}(\texttt{cream\_cheese},\texttt{basket})$
      & 0.0 & 0.0 & 10.0 & 0.0 & 0.0 & 0.0 & 12.0 & 6.0 & 98.0 & 99.0 & 100.0 & 100.0 \\
    $\operatorname{Place}(\texttt{ketchup},\texttt{basket})$
      & 0.0 & 0.0 & 0.0 & 0.0 & 20.0 & 2.0 & 32.0 & 12.0 & 96.0 & 100.0 & 90.0 & 100.0 \\
    $\operatorname{Place}(\texttt{milk},\texttt{basket})$
      & 0.0 & 0.0 & 0.0 & 0.0 & 0.0 & 0.0 & 38.0 & 2.0 & 98.0 & 98.0 & 100.0 & 100.0 \\
    $\operatorname{Place}(\texttt{orange\_juice},\texttt{basket})$
      & 0.0 & 0.0 & 0.0 & 0.0 & 0.0 & 2.0 & 30.0 & 2.0 & 100.0 & 100.0 & 100.0 & 100.0 \\
    $\operatorname{Place}(\texttt{salad\_dressing},\texttt{basket})$
      & 0.0 & 0.0 & 10.0 & 0.0 & 0.0 & 0.0 & 32.0 & 0.0 & 100.0 & 99.0 & 100.0 & 100.0 \\
    $\operatorname{Place}(\texttt{tomato\_sauce},\texttt{basket})$
      & 0.0 & 0.0 & 0.0 & 0.0 & 0.0 & 0.0 & 16.0 & 48.0 & 88.0 & 100.0 & 100.0 & 100.0 \\
    \midrule
    \textbf{Average}
      & 0.0 & 0.0 & 0.0 & 0.0 & 17.0 & 1.0 & 21.8 & 18.2
      & \textbf{96.0} & \textbf{98.0} & \textbf{99.0} & \textbf{100.0} \\
    \bottomrule
  \end{tabular}%
  }
\end{table}

\begin{table}[t]
  \centering
    \caption{Task-wise LIBERO-PRO performance on the \texttt{libero-goal}
    benchmark. Each entry reports position/task success rate in percentage (\%).
    $\dagger$ denotes the use of privileged simulator object poses.
    Results for OpenVLA, $\pi_0$, $\pi_{0.5}$, and CaP-Agent0 are taken
    directly from the CaP-Agent0 report~\citep{fu2026capx}.}
  \label{tab:libero-pro-goal-taskwise}
  \setlength{\tabcolsep}{3.0pt}
  \renewcommand{\arraystretch}{1.08}
  \resizebox{\textwidth}{!}{%
  \begin{tabular}{l*{12}{c}}
    \toprule
    \textbf{Task (Symbolic Form)}
      & \multicolumn{2}{c}{\textbf{OpenVLA}}
      & \multicolumn{2}{c}{$\boldsymbol{\pi_0}$}
      & \multicolumn{2}{c}{$\boldsymbol{\pi_{0.5}}$}
      & \multicolumn{2}{c}{\textbf{CaP-Agent0}}
      & \multicolumn{2}{c}{\textbf{RoboFoundry}}
      & \multicolumn{2}{c}{\textbf{RoboFoundry}$^{\boldsymbol{\dagger}}$} \\
    \cmidrule(lr){2-3}\cmidrule(lr){4-5}\cmidrule(lr){6-7}
    \cmidrule(lr){8-9}\cmidrule(lr){10-11}\cmidrule(lr){12-13}
      & Pos & Task & Pos & Task & Pos & Task & Pos & Task
      & Pos & Task & Pos & Task \\
    \midrule

    $\operatorname{Open}(\texttt{cabinet},\texttt{drawer\_mid})$
      & 0.0 & 0.0 & 0.0 & 0.0 & 0.0 & 4.0 & 0.0 & 0.0 & 71.0 & 71.0 & 95.0 & 100.0 \\
    $\operatorname{Put}(\texttt{bowl},\texttt{drawer\_top})$
      & 0.0 & 0.0 & 0.0 & 0.0 & 94.0 & 2.0 & 4.0 & 0.0 & 90.0 & 71.0 & 95.0 & 100.0 \\
    $\operatorname{Push}(\texttt{plate},\texttt{stove\_front})$
      & 0.0 & 0.0 & 0.0 & 0.0 & 0.0 & 0.0 & 0.0 & 10.0 & 71.0 & 70.0 & 95.0 & 100.0 \\
    $\operatorname{Put}(\texttt{bowl},\texttt{plate})$
      & 0.0 & 0.0 & 0.0 & 0.0 & 0.0 & 2.0 & 36.0 & 38.0 & 90.0 & 97.0 & 100.0 & 100.0 \\
    $\operatorname{Put}(\texttt{bowl},\texttt{stove})$
      & 0.0 & 0.0 & 0.0 & 0.0 & 0.0 & 4.0 & 22.0 & 12.0 & 99.0 & 100.0 & 95.0 & 100.0 \\
    $\operatorname{Put}(\texttt{bowl},\texttt{cabinet\_top})$
      & 0.0 & 0.0 & 0.0 & 0.0 & 0.0 & 2.0 & 60.0 & 4.0 & 100.0 & 99.0 & 99.0 & 100.0 \\
    $\operatorname{Put}(\texttt{cream\_cheese},\texttt{bowl})$
      & 0.0 & 0.0 & 0.0 & 0.0 & 98.0 & 2.0 & 4.0 & 34.0 & 98.0 & 90.0 & 94.0 & 100.0 \\
    $\operatorname{Put}(\texttt{wine\_bottle},\texttt{rack})$
      & 0.0 & 0.0 & 0.0 & 0.0 & 88.0 & 2.0 & 2.0 & 12.0 & 71.0 & 70.0 & 94.0 & 100.0 \\
    $\operatorname{Put}(\texttt{wine\_bottle},\texttt{cabinet\_top})$
      & 0.0 & 0.0 & 0.0 & 0.0 & 98.0 & 2.0 & 62.0 & 40.0 & 95.0 & 95.0 & 94.0 & 100.0 \\
    $\operatorname{TurnOn}(\texttt{stove})$
      & 0.0 & 0.0 & 0.0 & 0.0 & 0.0 & 0.0 & 66.0 & 18.0 & 95.0 & 97.0 & 99.0 & 100.0 \\
    \midrule
    \textbf{Average}
      & 0.0 & 0.0 & 0.0 & 0.0 & 38.0 & 0.0 & 25.6 & 16.8
      & \textbf{88.0} & \textbf{86.0} & \textbf{96.0} & \textbf{100.0} \\
    \bottomrule
  \end{tabular}%
  }
\end{table}

\begin{table}[htbp]
  \centering
  \caption{Task-wise LIBERO-PRO performance on the \texttt{libero-spatial}
  benchmark. Each entry reports position/task success rate in percentage (\%). $\dagger$
  denotes the use of privileged simulator object poses. Results for OpenVLA, $\pi_0$, $\pi_{0.5}$, and CaP-Agent0 are taken
directly from the CaP-Agent0 report~\citep{fu2026capx}.
  }
  \label{tab:libero-pro-spatial-taskwise}
  \setlength{\tabcolsep}{3.0pt}
  \renewcommand{\arraystretch}{1.08}
  \resizebox{\textwidth}{!}{%
  \begin{tabular}{l*{12}{c}}
    \toprule
    \textbf{Task (Symbolic Form)}
      & \multicolumn{2}{c}{\textbf{OpenVLA}}
      & \multicolumn{2}{c}{$\boldsymbol{\pi_0}$}
      & \multicolumn{2}{c}{$\boldsymbol{\pi_{0.5}}$}
      & \multicolumn{2}{c}{\textbf{CaP-Agent0}}
      & \multicolumn{2}{c}{\textbf{RoboFoundry}}
      & \multicolumn{2}{c}{\textbf{RoboFoundry}$^{\boldsymbol{\dagger}}$} \\
    \cmidrule(lr){2-3}\cmidrule(lr){4-5}\cmidrule(lr){6-7}
    \cmidrule(lr){8-9}\cmidrule(lr){10-11}\cmidrule(lr){12-13}
      & Pos & Task & Pos & Task & Pos & Task & Pos & Task
      & Pos & Task & Pos & Task \\
    \midrule

    $\operatorname{Pick}(\operatorname{between}(\texttt{plate},\texttt{ramekin}),\texttt{plate})$
      & 0.0 & 0.0 & 0.0 & 0.0 & 2.0 & 0.0 & 22.0 & 14.0 & 98.0 & 88.0 & 100.0 & 100.0 \\
    $\operatorname{Pick}(\texttt{table\_center},\texttt{plate})$
      & 0.0 & 0.0 & 0.0 & 0.0 & 0.0 & 2.0 & 22.0 & 14.0 & 89.0 & 99.0 & 100.0 & 100.0 \\
    $\operatorname{Pick}(\operatorname{drawer\_top}(\texttt{cabinet\_wood}),\texttt{plate})$
      & 0.0 & 0.0 & 0.0 & 0.0 & 0.0 & 0.0 & 2.0 & 10.0 & 88.0 & 96.0 & 99.0 & 98.0 \\
    $\operatorname{Pick}(\operatorname{next\_to}(\texttt{cookie\_box}),\texttt{plate})$
      & 0.0 & 0.0 & 0.0 & 0.0 & 0.0 & 2.0 & 0.0 & 10.0 & 97.0 & 96.0 & 100.0 & 98.0 \\
    $\operatorname{Pick}(\operatorname{next\_to}(\texttt{plate}),\texttt{plate})$
      & 0.0 & 0.0 & 0.0 & 0.0 & 0.0 & 0.0 & 10.0 & 20.0 & 97.0 & 88.0 & 100.0 & 98.0 \\
    $\operatorname{Pick}(\operatorname{next\_to}(\texttt{ramekin}),\texttt{plate})$
      & 0.0 & 0.0 & 0.0 & 0.0 & 12.0 & 2.0 & 30.0 & 14.0 & 89.0 & 96.0 & 85.0 & 100.0 \\
    $\operatorname{Pick}(\operatorname{on}(\texttt{cookie\_box}),\texttt{plate})$
      & 0.0 & 0.0 & 0.0 & 0.0 & 0.0 & 0.0 & 14.0 & 8.0 & 88.0 & 88.0 & 98.0 & 100.0 \\
    $\operatorname{Pick}(\operatorname{on}(\texttt{ramekin}),\texttt{plate})$
      & 0.0 & 0.0 & 0.0 & 0.0 & 98.0 & 2.0 & 2.0 & 20.0 & 98.0 & 88.0 & 99.0 & 100.0 \\
    $\operatorname{Pick}(\operatorname{on}(\texttt{stove}),\texttt{plate})$
      & 0.0 & 0.0 & 0.0 & 0.0 & 2.0 & 0.0 & 8.0 & 14.0 & 88.0 & 88.0 & 100.0 & 100.0 \\
    $\operatorname{Pick}(\operatorname{on}(\texttt{cabinet\_wood}),\texttt{plate})$
      & 0.0 & 0.0 & 0.0 & 0.0 & 90.0 & 0.0 & 8.0 & 16.0 & 88.0 & 88.0 & 99.0 & 98.0 \\
    \midrule
    \textbf{Average}
      & 0.0 & 0.0 & 0.0 & 0.0 & 20.0 & 1.0 & 11.8 & 14.0
      & \textbf{92.0} & \textbf{91.5} & \textbf{98.0} & \textbf{99.2} \\
    \bottomrule
  \end{tabular}%
  }
\end{table}

\end{document}